# Solved in Unit Domain:
# JacobiNet for Differentiable Coordinate-Transformed PINNs

Xi Chen[1], Jianchuan Yang[4], Junjie Zhang[5], Runnan Yang[6], Xu Liu[1], Hong Wang[1],
Tinghui Zheng[2]†, Ziyu Ren[3]†, Wenqi Hu[1]†

1. Department of Mechanical and Aerospace Engineering, The Hong Kong University of Science and Technology, Clear Water Bay, Kowloon, Hong Kong, China.
2. Department of Mechanics & Engineering, College Architecture & Environment, Sichuan University, Chengdu, China
3. School of Mechanical Engineering and Automation, Beihang University, Beijing 100191, China
4. Department of Civil and Environmental Engineering, The Hong Kong University of Science and Technology, Kowloon, Hong Kong, China.
5. College of Computing and Data Science, Nanyang Technological University, Singapore
6. West China Biomedical Big Data Center, West China Hospital, Sichuan University, Chengdu, 610041, China

## Abstract:

Physics-Informed Neural Networks (PINNs) offer a powerful framework for solving PDEs by embedding physical laws into the learning process. However, when applied to domains with irregular boundaries, PINNs often suffer from instability and slow convergence, which stems from (1) inconsistent normalization due to geometric anisotropy, (2) inaccurate boundary enforcement, and (3) imbalanced loss term competition. A common workaround is to map the domain to a regular space. Yet, conventional mapping methods rely on case-specific meshes, define Jacobians at pre-specified fixed nodes, reformulate PDEs via the chain rule—making them incompatible with modern automatic differentiation, tensor-based frameworks. To bridge this gap, we propose JacobiNet, a learning-based coordinate-transformed PINN framework that unifies domain mapping and PDE solving within an end-to-end differentiable architecture. JacobiNet enables direct Jacobian computation via autograd, shares computation graph with downstream PINNs, thereby avoiding case-specific meshing, explicit Jacobian computation/storage, and manual PDE reformulation while unlocking geometric-editing operations. Separating physical modeling from geometric complexity, JacobiNet (1) addresses normalization challenges in the original anisotropic coordinates, (2) facilitates the hard enforcement of boundary conditions, and (3) mitigates the long-standing imbalance among loss terms. Evaluated on various PDEs, JacobiNet reduces the $L_{2,\mathrm{rel}}$ error from 0.11–0.73 to 0.01–0.09, achieving an average 15.6× improvement in accuracy. In vessel-like domains with varying shapes, JacobiNet enables millisecond-level mapping inference for unseen geometries, improves prediction accuracy by an average of 3.65×, while delivering over 10× speedup—demonstrating strong generalization, accuracy, and efficiency.

Codes and datasets for this work will be made publicly available upon publication at https://github.com/xchenim/JacobiNet.

## 1. Introduction

By fusing physical laws with observation data, Physics-Informed Neural Networks (PINNs) offer a mesh-free paradigm for solving partial differential equations (PDEs) [1-3]. In this framework, a neural network takes spatial–temporal coordinates as inputs and predicts the corresponding physical quantities, with governing equations enforced via physics-based residuals computed through automatic differentiation. This recasts PDE solving as an optimization problem, enabling the direct incorporation of physical constraints into learning, thereby alleviating the reliance on large, labeled datasets and reducing the need for extensive data-driven training. Since inception, PINNs have achieved notable success across a wide range of application fields, including computational mechanics [4-6], materials science [7, 8], among others. Their differentiable formulation further supports inverse problems [1, 9], parameter identification [10-12], and PDE-constrained optimization [13, 14].

### 1.1 Mathematical formulation of physics-informed neural networks

Specifically, consider a general PDE system:

$$\mathcal{F}[u;\mathbf{x}] = 0, \mathbf{x} \in \Omega_{\mathbf{x}}, \tag{1}$$

$$\mathcal{B}[u] = 0, \mathbf{x} \in \partial\Omega_{\mathbf{x}}, \tag{2}$$

where $\mathcal{F}[\cdot]$ and $\mathcal{B}[\cdot]$ are the PDE residual and boundary operators, respectively, with $\mathcal{B}[\cdot]$ encompassing both spatial (e.g., Dirichlet, Neumann) and temporal (initial) conditions. $\Omega_{\mathbf{x}}$ is the physical domain of interest with boundaries $\partial\Omega_{\mathbf{x}}$. The network is typically constructed as a fully connected feedforward model $\mathcal{N}^L: \mathbb{R}^{D_i} \to \mathbb{R}^{D_o}$, where $D_i$ and $D_o$ denote the input and output dimensions of the network, respectively. The $k$-th hidden layer is defined as:

$$\mathcal{N}^k(\mathbf{x}) = \Phi^k\left(W_k \mathcal{N}^{k-1}(\mathbf{x}) + b_k\right),\ 1 \le k \le L-1, \tag{3}$$

with $\Phi(\cdot)$ denoting the activation function, $\theta = \{W_k, b_k\}$ the set of trainable parameters, and $\mathcal{N}^0(\mathbf{x}) = \mathbf{x} \in \mathbb{R}^{D_i}$ denoting the input layer.

The training process minimizes a composite loss function [1] that combines contributions from the PDE residuals, boundary conditions, and optionally observed data:

$$\mathcal{L}_{total} = \sum_{j=1}^{M_u} \lambda_{u_j} \mathcal{L}_{u_j} + \sum_{k=1}^{M_b} \lambda_{b_k} \mathcal{L}_{b_k} + \lambda_d \mathcal{L}_d, \tag{4}$$

where each component is defined as

$$\mathcal{L}_{u_j} = \frac{1}{N_{u_j}} \sum_{i=1}^{N_{u_j}} \left\| \mathcal{F}_j\left[\hat{u}\left(\mathbf{x}_i^{u_j}\right)\right] \right\|^2, \tag{5}$$

$$\mathcal{L}_{b_k} = \frac{1}{N_{b_k}} \sum_{i=1}^{N_{b_k}} \left\| \mathcal{B}_k\left[\hat{u}\left(\mathbf{x}_i^{b_k}\right)\right] \right\|^2, \tag{6}$$

$$\mathcal{L}_d = \frac{1}{N_d} \sum_{i=1}^{N_d} \left\| \hat{u}\left(\mathbf{x}_i^d\right) - u\left(\mathbf{x}_i^d\right) \right\|^2. \tag{7}$$

Here, $\mathcal{L}_{u_j}$, $\mathcal{L}_{b_k}$, and $\mathcal{L}_d$ denote the loss terms associated with the $j$-th PDE residual, the $k$-th boundary condition, and the optional data supervision, evaluated at the collocation points $\{x_i^{u_j}\}_{i=1}^{N_{u_j}}$, $\{x_i^{b_k}\}_{i=1}^{N_{b_k}}$, and $\{x_i^{d}\}_{i=1}^{N_d}$, respectively. The integers $M_u$ and $M_b$ denote the numbers of governing PDE operators and boundary-condition operators. The variable $u$ denotes the exact solution (or ground truth obtained from measurements or high-fidelity numerical simulations), while $\hat{u}$ represents the neural-network approximation. The weighting coefficients $\lambda_{u_j}$, $\lambda_{b_k}$, and $\lambda_d$ control the relative contribution of each loss term during training and may vary depending on the problem configuration.

### 1.2 Challenges of PINNs in irregular computational domains

While effective in simple domains with low geometric complexity, PINNs often suffer from instability and slow convergence when applied to domains with irregular boundaries.

First, geometric anisotropy undermines standard normalization strategies [15, 16]. Unlike scale-invariant numerical solvers like the finite volume method (FVM), PINNs are sensitive to input scales—improper normalization can disrupt activations, hinder gradient flow, and lead to training failure. Standard normalizations or nondimensionalizations apply only global stretching or compression along the original coordinate axes, typically limited to directions $x, y, z$. This rigid, global min-max operation fails to adapt to local geometric variations, reducing its effectiveness in handling complex, anisotropic structures. For example, in slender domains like 2D coronary arteries, simple normalization of $(x, y)$ fails to reconcile the large disparity between axial and radial scales, making it difficult to maintain consistent learning precision across all local directions $(r, z)$.

Second, the geometric complexity of irregular domains makes boundary condition sampling and enforcement more difficult. This, in turn, further amplifies training instabilities such as gradient degradation [17, 18], oscillatory convergence [19], and entrapment in suboptimal local minima [20, 21].

Finally, the PINN loss function comprises PDE residuals, boundary condition losses, and optional data supervision, which often differ in scale, gradient magnitude, and convergence behavior. This mismatch leads to imbalanced competition during optimization [18, 20, 22, 23]. The long-standing imbalance is further intensified in complex domains—even in unsupervised settings—models tend to overfit one component at the expense of others, degrading physical consistency and overall accuracy, as detailed in Appendix-1. Several strategies, such as gradient normalization [24], adaptive loss weighting [18, 22, 25], and residual-based attention mechanisms [26], have been proposed to alleviate this imbalance. However, these approaches primarily redistribute optimization effort, while the underlying causes of the imbalance remain. An alternative is to impose boundary conditions through hard-constraint trial functions [13, 20, 27]. However, for complex geometries with irregular boundaries, this approach can be difficult to apply in practice, as constructing explicit distance functions is often non-trivial.

In summary, these three challenges—as illustrated in Fig. 1A—underscore the limitations of applying standard PINNs directly to complex domains.

### 1.3 Coordinate transformation for complex domains

An empirically effective strategy is to map the irregular physical domain $\Omega_{\mathbf{x}}$ to a regular reference domain $\Omega_{\boldsymbol{\xi}}$ via coordinate transformation [28-30]. Through this mapping, spatial derivatives in the physical space are expressed in terms of reference-domain derivatives via the chain rule (detailed derivations are provided in Appendix-2), with the Jacobian $\boldsymbol{\mathcal{J}} = \partial\boldsymbol{\xi}/\partial\mathbf{x}$ and its derivatives governing the transformation. Consequently, the original PDEs and boundary conditions are reformulated in the reference domain, leading to the loss terms for PINN training:

$$\mathcal{L}_u = \sum_{j=1}^{M_u} \lambda_{u_j} \left(\frac{1}{N_{u_j}} \sum_{i=1}^{N_{u_j}} \left\| \mathcal{F}_j' \left[ \hat{u}\left(\boldsymbol{\xi}_i^{u_j}\right), \nabla_{\boldsymbol{\xi}} \hat{u}, \nabla_{\boldsymbol{\xi}}^2 \hat{u}, \boldsymbol{\mathcal{J}}_i \right] \right\|^2 \right), \tag{8}$$

$$\mathcal{L}_b = \sum_{k=1}^{M_b} \lambda_{b_k} \left(\frac{1}{N_{b_k}} \sum_{i=1}^{N_{b_k}} \left\| \mathcal{B}_k' \left[ \hat{u}\left(\boldsymbol{\xi}_i^{b_k}\right), \nabla_{\boldsymbol{\xi}} \hat{u}, \nabla_{\boldsymbol{\xi}}^2 \hat{u}, \boldsymbol{\mathcal{J}}_i \right] \right\|^2 \right), \tag{9}$$

where the operators $\mathcal{F}_j'[\cdot]$ and $\mathcal{B}_k'[\cdot]$ are now applied on reference-domain predictions, transformed back into physical space using the $\boldsymbol{\mathcal{J}}_i$, the Jacobian matrix of the mapping evaluated at the $i$-th point $\mathbf{x}_i$. Thus, constructing a bijective and smooth mapping $\Phi: \mathbf{x} \mapsto \boldsymbol{\xi}$ with accessible Jacobians is crucial for this strategy.

In conventional numerical methods and mesh generation, various mapping strategies have been developed, including analytic mappings, curvilinear coordinate transformations [27, 31, 32], and diffeomorphic transformations [29] (PDE-based elliptic mappings [31, 32]). These methods can, to some extent, regularize complex physical domains into regular reference domains to facilitate solution procedures, while they remain hindered by several limitations. First, these methods are case-specific: any change in geometry requires recomputing the entire mapping, preventing generalization to new domains. Second, approaches like PDE-based elliptic mappings require prior mesh generation, which conflicts with the mesh-free nature emphasized by PINNs. Dense meshes are often required to ensure the positive definiteness and robustness of the Jacobian, thereby inflating preprocessing cost and computational overhead. Moreover, mapping information is typically defined only at discrete nodes. Evaluating Jacobians at arbitrary sampling points requires interpolation, which introduces errors and disrupts continuity. Then the Jacobian matrix itself—along with its determinant and inverse—must be explicitly stored and manually incorporated into the reformulated PDEs via the chain rule. This non-differentiable construction process cannot be directly integrated into the PINN training workflow, nor does it support gradient-based or residual-based optimization [18, 22, 33-35]. In addition, existing methods are sensitive to boundary continuity and smoothness, and tend to fail in the presence of sharp corners, discontinuities, or local noise, or more complex domains such as those with multiple (>5) $C^0$ continuous edges [28] or multiconnected structures. Finally, numerical Jacobian computation (e.g., finite differences [28]) introduces truncation errors, and potential local discontinuities. As summarized in Fig. 1B, these drawbacks render conventional mapping methods incompatible with the automatic differentiation, tensor-based workflows of modern PINN frameworks, preventing them from serving as a mainstream tool for complex domains in scientific computing.

In recent years, several studies have explored learning-based parameterization as a more flexible and differentiable alternative [36, 37]. Deep neural networks have been employed to achieve smooth parameterizations while reducing distortion [38]. Extensions from 2D to volumetric domains have further demonstrated the feasibility of obtaining high-quality and bijective mappings for complex geometries [39]. Other approaches incorporate boundary information [40] or leverage graph-based neural operators to enhance robustness and generalization [41], often achieving both higher efficiency and better adaptability compared with classical numerical mappings. However, these approaches are generally developed independently of the physics solver, leaving the geometric mapping and PDE solution decoupled. When applied to PINNs, the Jacobian is still excluded from the computational graph, leaving PDEs and boundary conditions to be manually reformulated via the chain rule—just as in conventional numerical methods. Recent coordinate transformation–based PINN methods exhibit similar limitations. For example, CT-PINN [42] employs a characteristic-based coordinate transformation to simplify solution structures for hyperbolic conservation laws with discontinuities, yet the governing equations must still be reformulated accordingly.

In summary, conventional numerical mappings are mesh-based, non-differentiable, and case-specific, while learning-based parameterizations remain decoupled from physics solvers. Consequently, these methods rely on pre-defined Jacobians and manual reformulation procedures, limiting their generality and adaptability.

### 1.4 Scope and contributions of present work

To address these challenges, we propose **JacobiNet**, a learning-based coordinate-transformed PINN framework that unifies domain mapping and PDE solving within an end-to-end differentiable architecture. JacobiNet learns a smooth mapping from irregular physical domains to a regular reference space, with Jacobians computed directly via automatic differentiation, avoiding mesh generation and reducing the need for manual PDE reformulation. By decoupling geometric complexity from physical modeling, JacobiNet improves normalization consistency, facilitates accurate boundary enforcement, and alleviates loss-term imbalance in irregular domains.

The structure of this paper is as follows: **Section 2** introduces JacobiNet, detailing its architecture, comparisons with existing mappings (Fig. 2A), geometric-editing capabilities (Fig. 2B), and end-to-end integration with PDE solving (Fig. 3A–C). **Section 3** presents numerical experiments, ablation studies, and computational overhead analysis, comparing JacobiNet with baseline and the state-of-the-art methods (Fig. 4-7) on several benchmark PDE problems. **Section 4** concludes the paper and discusses drawbacks and future directions.

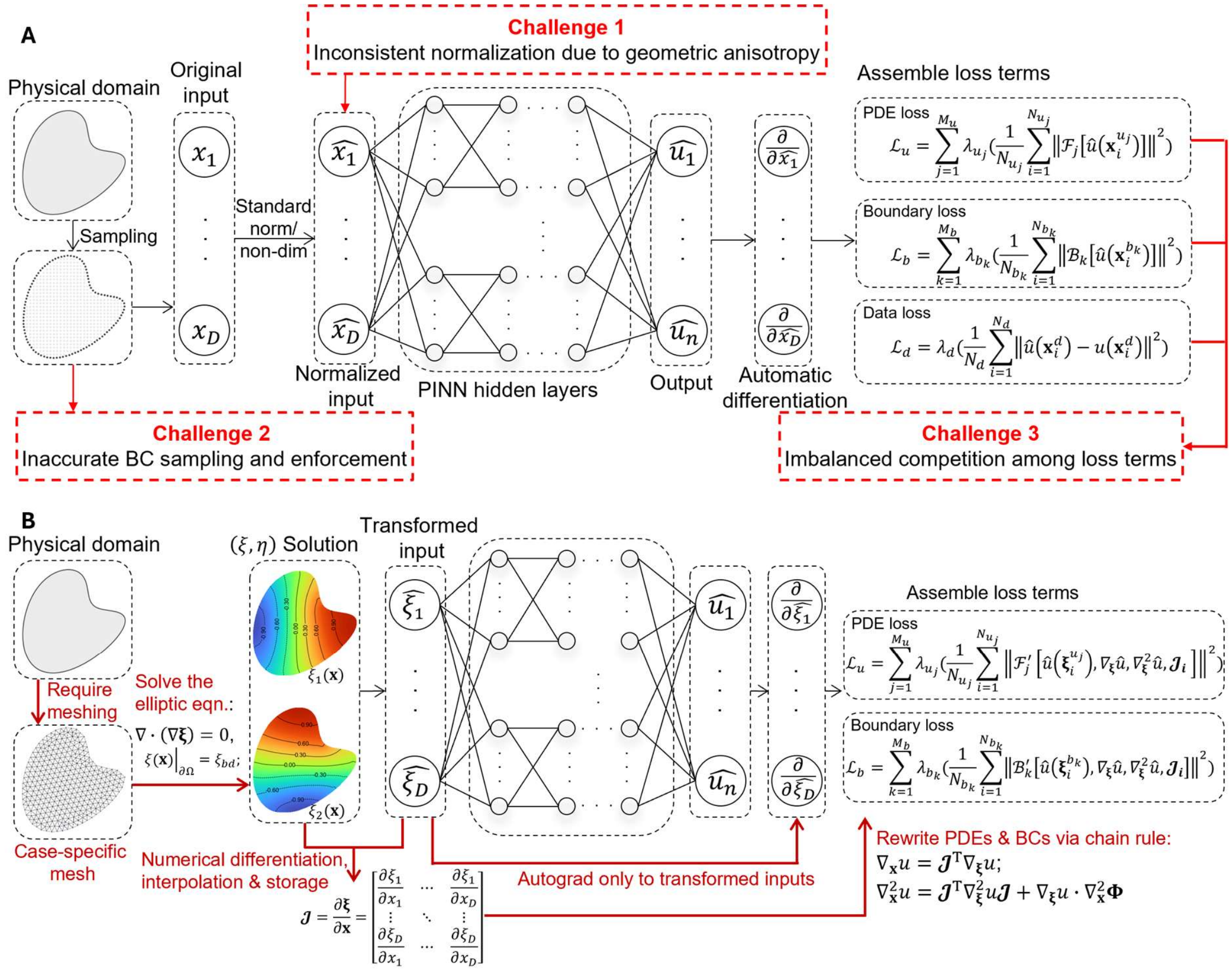

**Figure 1. Standard & coordinate-transformed PINN workflows in complex domains and challenges. (A)** Standard PINN workflow and three key challenges in complex domains. **(B)** Conventional coordinate-transformed PINNs workflow and limitations. The process begins with case-specific meshing of the physical domain, followed by numerically solving elliptic PDEs to obtain a bijective mapping $\mathbf{x} \mapsto \boldsymbol{\xi}$. The mapped coordinates $\boldsymbol{\xi}$ serve as PINN inputs, while the Jacobian matrix $\mathcal{J}$ is computed via numerical differentiation and interpolation. Since autograd operates only on $\boldsymbol{\xi}$, the PDE and boundary losses must be reformulated using the chain rule with the prestored Jacobian $\mathcal{J}$.

# 2. JacobiNet

To unify domain mapping and PDE solving in an end-to-end pipeline, we propose JacobiNet, a learning-based coordinate-transformed PINN framework. In the following, we detail its architecture, supervision strategy, loss formulation, as well as the way JacobiNet supports geometric editing operations and integrates geometric mappings seamlessly into the PINN framework.

## 2.1 Network architecture and loss function

JacobiNet uses lightweight multilayer perceptrons (MLP) with the hyperbolic tangent (tanh) activation function to construct a continuous, differentiable transformation $\Phi_{\vartheta}: \mathbf{x} \mapsto \boldsymbol{\xi}$, where $\mathbf{x} \in \Omega_{\mathbf{x}} \subset \mathbb{R}^D$ denotes the physical coordinates and $\boldsymbol{\xi} \in \Omega_{\boldsymbol{\xi}} \subset \mathbb{R}^D$ denotes the reference coordinates. Unlike numerical coordinate transformations that require mesh generation, JacobiNet learns the mapping through supervised point-pair training, without dependence on mesh.

The total loss function is defined as:

$$\mathcal{L} = \frac{1}{N_{in}}\sum_{i=1}^{N_{in}}\left\|\Phi_\vartheta\left(\mathbf{x}_i^{in}\right) - \boldsymbol{\xi}_i^{in}\right\|^2 + \lambda\frac{1}{N_{bd}}\sum_{i=1}^{N_{bd}}\left\|\Phi_\vartheta\left(\mathbf{x}_i^{bd}\right) - \boldsymbol{\xi}_i^{bd}\right\|^2, \tag{10}$$

here, the training set is split into internal points $\mathbf{x}_i^{in} \in \Omega$ and boundary points $\mathbf{x}_i^{bd} \in \partial\Omega$. $\lambda > 1$ strengthens the supervision on boundary points to improve the mapping accuracy at the boundary. In our work, we set $\lambda = 10$. The sensitivity analysis of $\lambda$, as well as its impact on boundary mapping accuracy and PDE solution accuracy, is provided in Appendix-3.

For each case, we monitor the root mean square error $RMSE_{in}$, $RMSE_{bd}$ and the maximum normal deviation $\mathcal{E}_{max}$ of boundary points, defined as:

$$RMSE_{in} = \left(\frac{1}{N_{in}}\sum_{i=1}^{N_{in}}\left\|\Phi_\vartheta\left(\mathbf{x}_i^{in}\right) - \boldsymbol{\xi}_i^{in}\right\|^2\right)^{\frac{1}{2}}, \tag{11}$$

$$RMSE_{bd} = \left(\frac{1}{N_{bd}}\sum_{i=1}^{N_{bd}}\left\|\Phi_\vartheta\left(\mathbf{x}_i^{bd}\right) - \boldsymbol{\xi}_i^{bd}\right\|^2\right)^{\frac{1}{2}}, \tag{12}$$

$$\mathcal{E}_{max} = max\left|n_{bd,i}\cdot\left[\Phi_\vartheta\left(\mathbf{x}_i^{bd}\right) - \boldsymbol{\xi}_i^{bd}\right]\right|, \tag{13}$$

here $n_{bd,i}$ denotes the outward unit normal of the corresponding edge. The $\mathcal{E}_{max}$ term quantifies the normal displacement (either inward or outward) from the edge, while tangential freedom is preserved, evaluating whether the mapped boundary points remain on the boundary.

Since JacobiNet is trained using paired point correspondences between the physical and reference domains, the supervision itself implicitly encourages mappings that preserve injectivity. In our experiments, explicitly enforcing the constraint $\det(\boldsymbol{\mathcal{J}}) > 0$ was found to complicate the optimization and degrade the fitting accuracy. Therefore, we use the Jacobian determinant only as a diagnostic metric rather than a training objective. To verify the local injectivity, we evaluated the positive Jacobian determinant ratio after training, defined as the proportion of sampled points whose Jacobian determinant is positive:

$$r_{\det(\mathcal{J})} = \frac{\#\{(\mathbf{x}_i)\in\Omega \mid \det(\mathcal{J}(\mathbf{x}_i))>0\}}{\#\{(\mathbf{x}_i)\in\Omega\}}, \tag{14}$$

where $\#\{\cdot\}$ is the cardinality (number of elements) of a set. $\boldsymbol{\mathcal{J}}(\mathbf{x}_i)$ is the Jacobian matrix of the transformation $\Phi_\vartheta$ at the $i$-th point $\mathbf{x}_i$, as defined by $\boldsymbol{\mathcal{J}}(\mathbf{x}_i) = \partial\boldsymbol{\xi}_i/\partial\mathbf{x}_i$. A positive determinant ($\det(\boldsymbol{\mathcal{J}}(\mathbf{x}_i)) > 0$) indicates that the mapping at that point is locally bijective and preserves orientation.

## 2.2 Support for geometric editing operations

JacobiNet computes Jacobian tensors directly via automatic differentiation. By propagating derivatives through the network's computational graph, automatic differentiation avoids truncation and discretization errors of numerical schemes, producing results that are mathematically exact up to machine limits ($\sim 10^{-16}$ in double precision). Combined with the smooth network parameterization used here, this yields continuous Jacobian fields in our implementation. Remarkably, a neural network-based mapping brings Jacobian access into geometric editing operations such as filling, radial stretching, and cut-and-unfold flattening (Fig. 2B)—operations that are not readily supported in conventional physics solvers, as their Jacobians are typically unavailable. Moreover, geometric editing operations, enabled by data supervision of JacobiNet, can handle complex domains—including those with more than five $C^0$-continuous boundaries or even some multiconnected structures—where applying numerical elliptic transformations would fail. By eliminating the need for explicit formulations, numerical mapping solvers, and mesh-based PDE reformulations, this neural network-based mapping also reduces the technical barrier and computational cost of coordinate transformation, making high-quality Jacobians accessible in geometries that were previously prohibitively expensive to handle.

A point-to-point comparison among conventional numerical mapping methods, learning-based approaches and JacobiNet is provided in Fig. 2A. In the examples shown in Fig. 2B, we demonstrate representative editing operations with their corresponding transformed coordinates, JacobiNet predictions, and the smoothness of the resulting Jacobians. The detailed algorithms of two representative geometry editing strategies used in this work (radial stretching & unfolding along centerline) are provided in Appendix-6. For all cases, uniform sampling was applied, and a standard 70% / 15% / 15% split was used for training, validation, and testing, respectively. Across these examples, JacobiNet learns diverse mappings for complex geometries while maintaining nearly fold-free and low-distortion properties. For the L-shaped and hexagram domains, the Jacobian determinant ratio reaches $r_{\det(\mathcal{J})} = 100\%$, indicating $\det(\boldsymbol{\mathcal{J}}(x_i, y_i)) > 0$ at all sampled points. For the concentric-ring domain, the ratio slightly decreases to 98.48%, with a small number of negative $\det(\boldsymbol{\mathcal{J}})$ values confined to two narrow regions aligned with the cut direction used to unfold the annular domain—one near the artificial seam and the other on the opposite side of the ring—where the mapping becomes locally more challenging. The prediction errors are low, with an average $RMSE_{in}/RMSE_{bd}/\mathcal{E}_{max}$ below 7.6/1.9/1.5 mm. Given that the characteristic size of the original geometries ranges from 1.5 m to 2 m, these errors correspond to relative magnitudes below 0.4% for all cases, demonstrating the high accuracy of the learned mapping. The visualized Jacobian determinants exhibit smooth and continuous variation, confirming the effectiveness of point-pair supervision. The training curves further show that such high-quality coordinate mappings and smooth Jacobians can be learned in as little as 45.3s on average.

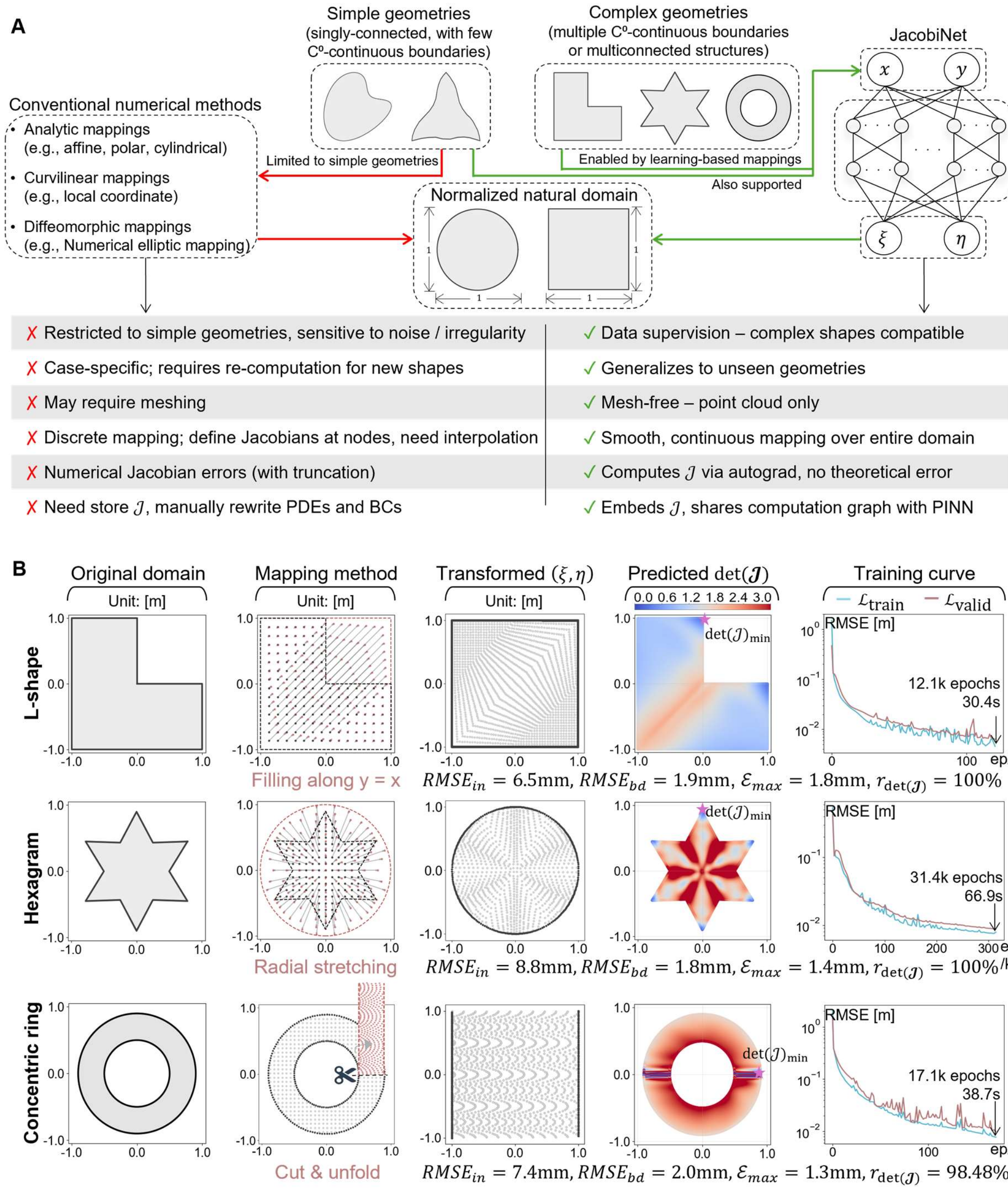

**Figure 2. JacobiNet enables continuous, differentiable mappings and geometric editing operations, surpassing the limitations of conventional numerical methods.** (**A**) Point-to-point comparisons between conventional numerical mapping methods and the proposed JacobiNet framework. (**B**) JacobiNet enables geometry editing through paired data supervision. From left to right: original domain, mapping method, transformed coordinates $(\xi, \eta)$, predicted Jacobian determinant $\det(\boldsymbol{\mathcal{J}})$, and training curve.

## 2.3 JacobiNet-PINN: An end-to-end differentiable framework of domain mapping and PDE solving

As shown in Fig. 3A, JacobiNet learns a differentiable coordinate mapping. It shares a unified computational graph with the downstream PINN solver, enabling seamless gradient propagation. Together, they form an end-to-end differentiable framework for domain mapping and PDE solving. In the transformed regular domain, Dirichlet boundary conditions are naturally imposed via trial functions, following the formulation of exact boundary-condition enforcement proposed in [27].

$$\hat{u}(\mathbf{x}) = P_u(\mathbf{x}) + D_u(\Phi_\vartheta(\mathbf{x})) \cdot \mathcal{N}^L(\mathbf{x}), \tag{15}$$

here, $\hat{u}(\mathbf{x})$ is the final approximation of the solution, $\mathcal{N}^L(\mathbf{x})$ denotes the raw output of the integrated pipeline, and $\Phi_\vartheta(\mathbf{x})$ represents the JacobiNet transformation. $P_u(\mathbf{x})$ is a prescribed function that satisfies the Dirichlet boundary conditions. The distance function $D_u(\Phi_\vartheta(\mathbf{x}))$ is defined to vanish on the boundary, thereby ensuring that the contribution from the neural network term is zero at the boundary. For example, in cases where $u(\mathbf{x}) = g(\mathbf{x})$ is required on a complex boundary, JacobiNet can flatten it to a simple, straight edge in the reference domain (e.g., $\Phi_\vartheta(\mathbf{x}_i^{bd}) = a$), allowing the hard imposition of boundary conditions to be formulated analytically. For example, let $P_u(\mathbf{x}) = g(\mathbf{x})$ and $D_u(\Phi_\vartheta(\mathbf{x})) = (a - \Phi_\vartheta(\mathbf{x}))$ or $(1 - e^{a-\Phi_\vartheta(\mathbf{x})})$.

We illustrate the method using the two-dimensional Poisson equation defined on a non-convex domain, approximately bounded within $[-1\text{m}, 1\text{m}]$ in both spatial directions. The governing equations are:

$$\nabla \cdot \big(\nabla u(\mathbf{x})\big) - f(\mathbf{x}) = 0, \mathbf{x} \in \Omega_p, \tag{16}$$

$$u(\mathbf{x}) = 0, \mathbf{x} \in \partial\Omega_p, \tag{17}$$

where $u(\mathbf{x})$ is the target scalar field, subject to zero Dirichlet boundary conditions on all edges. The source term [43] is defined as:

$$f(\mathbf{x}) = -2\pi^2 \sin(\pi X)\sin(\pi Y), \tag{18}$$

with $X$, $Y$ denoting the spatial coordinates.

In what follows, we compared performance among baseline PINN, JacobiNet-PINN and chain-rule-based coordinate-transformed PINN. All models are trained for 1,000 epochs under identical settings, including network architecture, optimizer configuration, and total number of trainable parameters. Specifically, JacobiNet employs two hidden layers with 128 neurons each and Tanh activations, while the PINN solver consists of three hidden layers with 64 neurons and SiLU activations. All models are optimized using the Adam optimizer with an initial learning rate of $10^{-3}$, followed by a cosine annealing schedule that decays the learning rate to $10^{-5}$.

In Fig. 3B, JacobiNet learns geometric editing operations (radial stretching) to transform irregular physical coordinates into a regular circular domain, achieving $RMSE_{in}/RMSE_{bd}/\mathcal{E}_{max} = 6.6/1.5/0.9$mm (0.33%/0.075%/0.045% of the transformed domain). Moreover, $r_{\det(\mathcal{J})} = 100\%$ further indicates the mapping's local injectivity and absence of fold-overs. Building on this, JacobiNet enables precise PINN predictions, reducing the relative error from $L_{2,\text{rel}} = 0.540$ (baseline PINN, Fig. 3C) to $L_{2,\text{rel}} = 0.015$, representing an improvement of over 10×.

Furthermore, compared with our framework, chain-rule-based approaches show reduced accuracy ($L_{2,\mathrm{rel}} > 0.031$), limited by numerical errors in Jacobian computation and by storage precision (double, single, or millimeter-level). In particular, the numerical errors mainly originate from numerical approximation of the Jacobian and its derivatives, where a second-order central finite difference scheme can introduce the discretization truncation error as $O(h^2)$, together with round-off accumulation, further amplifies inaccuracies in chain-rule reformulations. As shown in Fig. 3C, double precision (64-bit, ~15–16 significant digits), single precision (32-bit, ~6–7 digits), and millimeter-level precision (~3 decimal places) produce progressively degraded results. Even when using Jacobians obtained via automatic differentiation (as in JacobiNet), accuracy still degrades ($L_{2,\mathrm{rel}} = 0.031$) due to floating-point round-off. In contrast, our end-to-end framework embeds Jacobians directly within the computational graph, eliminating the need for numerical approximation or external storage, and thereby avoiding numerical and storage-induced errors. In addition, to ensure accurate first- and second-order derivatives in the PDE, the chain-rule expansion requires explicit inclusion of the Jacobian terms $\frac{\partial \xi}{\partial x}, \frac{\partial \xi}{\partial y}, \frac{\partial \eta}{\partial x}, \frac{\partial \eta}{\partial y}$, as well as the second-order terms $\frac{\partial^2 \xi}{\partial x^2}, \frac{\partial^2 \xi}{\partial y^2}, \frac{\partial^2 \xi}{\partial x \partial y}, \frac{\partial^2 \eta}{\partial x^2}, \frac{\partial^2 \eta}{\partial y^2}, \frac{\partial^2 \eta}{\partial x \partial y}$. This substantially increases the effective input dimensionality—for example, in 2D one must include 4 first-order terms and 6 second-order terms—resulting in a 5× increase, leading to significantly higher computational cost. In practice, their efficiency is much lower (>0.50 s/epoch) compared to our parallel tensorized end-to-end framework (0.015 s/epoch).

To further evaluate the robustness of the proposed JacobiNet under noisy conditions, we conducted experiments by injecting Gaussian noise into the normalized coordinates used for training. As detailed in Appendix-4, four noise levels were considered, namely 0%, 1%, 10%, and 30% of the sampling resolution. The results show that the learned mappings remain fold-free for noise levels up to 10%. Even at 30% noise, only a small local region exhibits negative Jacobian determinants, while the injectivity metric remains high at $r_{\det(\mathcal{J})} = 98.64\%$. In terms of the downstream PDE solution, the relative $L_2$ error of the Poisson problem increases with noise magnitude but remains within an acceptable range $L_{2,\mathrm{rel}} < 0.10$ for noise levels up to 10%. Even under 10% noise, the error of the predicted solution remains substantially lower than that of the standard baseline PINN, for which $L_{2,\mathrm{rel}} = 0.54$.

In summary, by separating physical modeling from geometric complexity, the seamless integration of geometric mapping and PDE solving, together with trial functions constructed in the mapped domains, efficiently addresses three key challenges of PINNs in complex domains—normalization in anisotropic coordinates, inaccurate boundary enforcement, and imbalanced loss-term competition. Consequently, JacobiNet achieves substantially better performance than baseline and chain-rule-based PINNs.

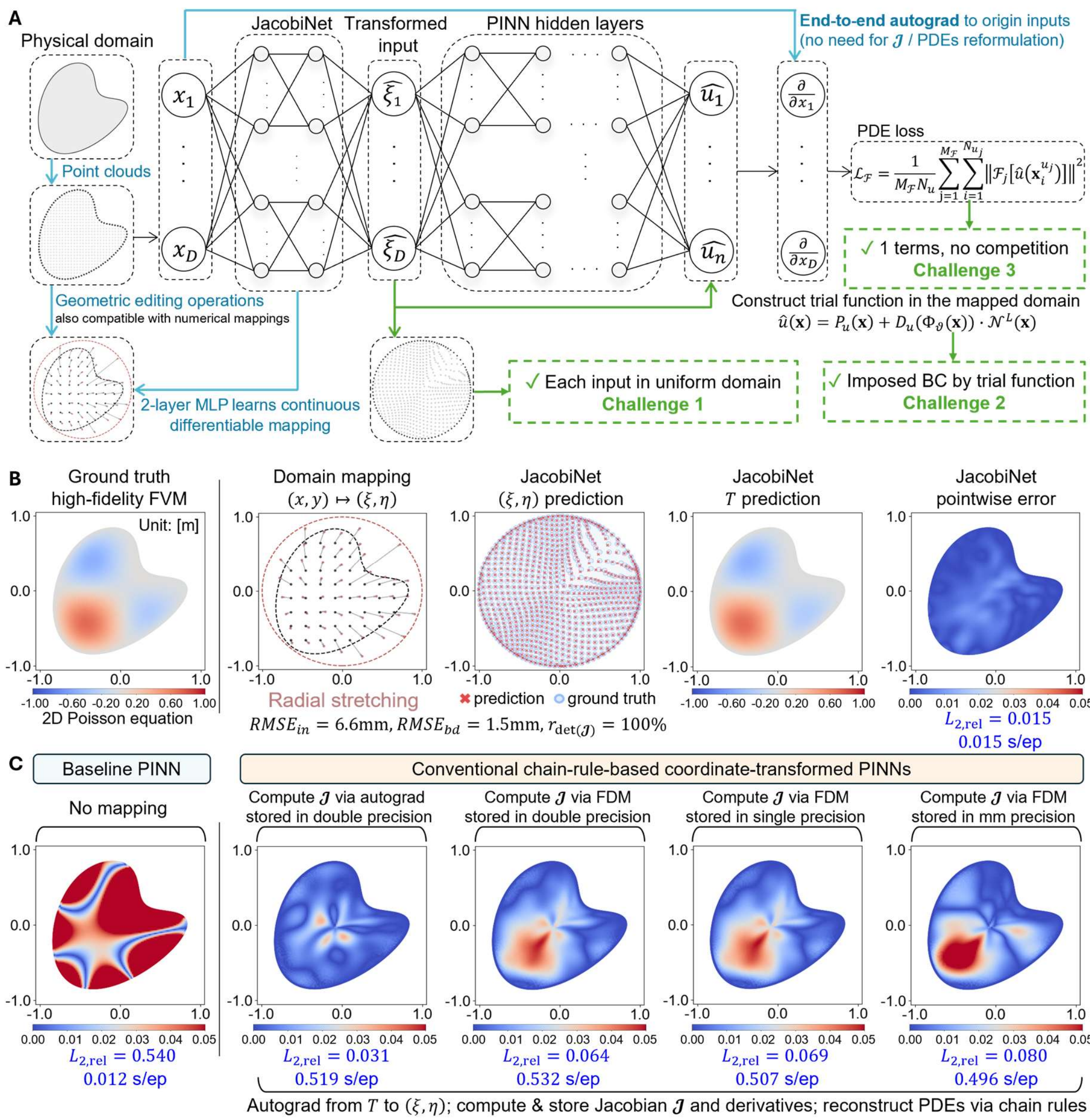

**Figure 3. JacobiNet's end-to-end differentiable integration of domain mapping and PDE solving.** **(A)** Workflow of JacobiNet + PINN. **(B)** Comparison between ground-truth FVM results and JacobiNet predictions. JacobiNet produces accurate transformed coordinates with $r_{\det(\mathcal{J})} = 100\%$, enabling PINN predictions with $L_{2,\text{rel}} = 0.015$, achieving an improvement of over 10×. **(C)** Comparison of accuracy and efficiency with baseline PINNs and conventional chain-rule–based transformed PINNs that require explicit PDE reconstruction using pre-computed and stored $\mathcal{J}$.

# 3. Numerical Experiments

To validate the effectiveness of the proposed frameworks, we applied JacobiNet to learn coordinate transformations for various irregular geometries, including U-shaped, S-shaped, T-shaped domains, three-dimensional stenosed vessel models, and vessel-like geometries with varying lengths and deformations. The corresponding Jacobian components were visualized via automatic differentiation, and the associated metrics ($RMSE_{in}$, $RMSE_{bd}$, $\mathcal{E}_{max}$ and $r_{\det(\mathcal{J})}$) were also reported, demonstrating the smooth and continuous nature of the learned coordinate transformations, as well as their local injectivity without fold-overs under various mapping operations.

In section 3.1-3.5, we applied JacobiNet to solve a range of representative PDEs, including the Laplace equation (Section 3.1), Poisson equation (Section 3.2), Helmholtz equation (Section 3.3), and the Navier–Stokes equations (Section 3.4, 3.5). In all experiments, solution accuracy was evaluated by the relative $L_2$ error, defined as:

$$L_{2,\text{rel}} = \frac{\sqrt{\sum_{i=1}^{N}[u(\mathbf{x}_i)-\hat{u}(\mathbf{x}_i)]^2}}{\sqrt{\sum_{i=1}^{N}[u(\mathbf{x}_i)]^2}}, \tag{19}$$

here, $u(\mathbf{x}_i)$ denotes the reference FVM solution computed on a high-resolution mesh, and $\hat{u}(\mathbf{x}_i)$ represents the model prediction.

In section 3.6, we present a computational overhead analysis to quantify the additional cost introduced by JacobiNet, showing that the added cost is moderate and offset by its efficiency—requiring on average up to an order of magnitude fewer epochs to reach the same accuracy (Section 3.7), thus substantially reducing total training time.

In section 3.7, we compare JacobiNet against a baseline PINN and several state-of-the-art (SOTA) methods. We further highlight the lightweight and plug-and-play nature of JacobiNet by demonstrating its effortless integration with existing approaches, such as a gradient-based reweighting scheme (Gradient Normalization, Grad Norm) [18] and a random Fourier feature (RFF) embedding strategy [44].

All experiments were conducted on a GeForce RTX 4090D GPU using PyTorch 2.4.1. JacobiNet is implemented as a fully connected neural network with 2 hidden layers, each containing 128 neurons and Tanh activations. The PINN employs 3 hidden layers in 2D cases (Section 3.1-3.3, 3.5-3.6) and 4 hidden layers in 3D cases (Section 3.4), each with 64 neurons and SiLU activations. The training process uses the Adam optimizer with an initial learning rate $10^{-3}$, and a Cosine Annealing Learning Rate Scheduler, which gradually reduces the learning rate to $10^{-5}$. Unless otherwise specified, all loss weighting coefficients $\lambda_i$ are set to 1. In each case, to ensure a fair comparison, all methods share identical network configurations, training setups, and evaluation criteria across all test cases, respectively. Further configuration details are provided in Appendix-5.

## 3.1 Laplace equation in U-shape

To evaluate the applicability of the proposed method on a representative scalar field problem, we consider the two-dimensional Laplace equation—a classical boundary value problem (BVP) that models steady-state phenomena:

$$\nabla \cdot \big(\nabla u(\mathbf{x})\big) = 0, x \in \Omega_p, \tag{20}$$

$$u(\mathbf{x}) = \pm 1, \mathbf{x} \in \partial\Omega_1, \mathbf{x} \in \partial\Omega_3, \tag{21}$$

$$\frac{\partial u(\mathbf{x})}{\partial n} = 0, \mathbf{x} \in \partial\Omega_2 \cup \partial\Omega_4, \tag{22}$$

as shown in Fig. 4A-1, $\Omega_p$ denotes the U-shaped physical domain, where $u(x)$ is the scalar field to be solved. The physical coordinates are defined within the range $x, y \in [0,1]$ cm. Dirichlet boundary conditions are prescribed as $u = \pm 1$ on the lower and upper boundaries $\partial\Omega_1$ and $\partial\Omega_3$, respectively. Neumann conditions are applied to the lateral walls $\partial\Omega_2$ and $\partial\Omega_4$, enforcing zero normal derivatives. In this case, while the Dirichlet conditions on $\partial\Omega_1$ and $\partial\Omega_3$ are imposed via a trial function, Neumann boundary conditions are enforced by adding a boundary loss term to the objective. Directly embedding derivative constraints would otherwise require the network to explicitly predict both the solution and its gradients, along with introducing additional consistency loss terms, which lies beyond the scope of this work.

We begin by unfolding the physical domain (in cm) along its centerline and normalizing it to a regular reference domain (in m). JacobiNet is then trained to learn the mapping from physical coordinates $(x, y)$ to reference coordinates $(\xi, \eta)$. The mapping achieves high accuracy, with quantitative metrics of $RMSE_{in}$ = 2.6mm, $RMSE_{bd}$ = 1.8mm, $\mathcal{E}_{max}$ = 5.0mm, $r_{\det(\mathcal{J})} = 100\%$. To further illustrate the smoothness of the learned transformation, we examine the lower boundary $\partial\Omega_1$ as an example to plot the directional derivatives of the predicted $\nabla\xi$ and $\nabla\eta$ along the normalized arc length (which also corresponds to $\eta$ in this case). Both $\mathcal{J}$ components exhibit smooth, periodic variation consistent with the underlying geometry, without abrupt jumps.

Fig. 4B-1 represents the FVM solution on a high-fidelity grid, the corresponding representation in the transformed domain. Fig. 4C-1 presents the pointwise error of baseline PINN's and JacobiNet's prediction. Compared with the baseline PINN trained directly in the physical domain, the JacobiNet reduces the relative $L_2$ error from 0.245 to 0.033.

### 3.2 Poisson equation in S-shape

In this case, we consider a two-dimensional Poisson equation defined over an S-shaped domain (Fig. 4A-2). The domain exhibits pronounced curvature, which presents challenges for standard PINNs, particularly in normalization, where uniform scaling in Cartesian coordinates fails to preserve both axial and radial consistency. The governing equations are given by Eqs. (16)-(18).

JacobiNet learns a mapping pattern similar to that in the U-shaped case, unfolding the S-shaped domain into a regular reference space. Quantitatively, the mapping achieves $RMSE_{in}$, $RMSE_{bd}$, $\mathcal{E}_{max}$ equal to 3.4mm, 1.8mm, 5.7mm, with $r_{\det(\mathcal{J})} = 100\%$. The mapping accuracy is slightly lower than in the U-shaped case due to stronger curvature and larger boundary variations in the S-shaped geometry.

In this case, JacobiNet achieves uniformly lower pointwise errors and a relative $L_2$ error of 0.015, compared to 0.319 for the baseline PINN using standard normalization—an improvement of over 20×.

### 3.3 Helmholtz equation in T-shape

To further evaluate JacobiNet's capability in handling geometric anisotropy and PDEs with high-frequency behavior, we consider a Helmholtz equation defined over a T-shaped domain (Fig. 4A-3), featuring a narrow throat connecting two wide chambers. The governing equations are:

$$\nabla \cdot \left(\nabla u(\mathbf{x})\right) + k^2 u(\mathbf{x}) - f(\mathbf{x}) = 0, \mathbf{x} \in \Omega_p, \tag{23}$$

$$u(\mathbf{x}) = 0, \mathbf{x} \in \partial\Omega_p. \tag{24}$$

The source term is:

$$f(\mathbf{x}) = (k^2 - (a_1\pi)^2 - (a_2\pi)^2)\sin(a_1\pi X)\sin(a_2\pi Y), \tag{25}$$

where $k = 1$, $a_1 = 2$, $a_2 = 6$ chosen to emphasize high-frequency behavior, $X$ and $Y$ are the spatial coordinates.

Here we apply a stretching operation—specifically, a Y-slice min–max normalization—that laterally stretches each y-slice. The narrow throat region $-0.1 \le x \le 0.1$ is expanded to $-1 \le \xi \le 1$, which can be denoted as $\xi = \mathrm{norm}(x)$. In this case, the mapping accuracy is $RMSE_{in} = 7.3$mm, $RMSE_{bd} = 1.4$mm, $\mathcal{E}_{max} = 7.3$mm, and $r_{\det(\mathcal{J})} = 100\%$. To further analyze the transformation, we examine the bottleneck line at $y = 0.3$cm and plot the corresponding Jacobian components. As expected, $\partial\eta/\partial x = 0$ and $\partial\eta/\partial y = 200$, indicating a pure x-direction rescaling where $\eta = y$. Meanwhile, $\partial\xi/\partial x$ varies smoothly from 2000 to –2000 across the stretched region, consistent with the 2000× lateral expansion applied at the throat (from [-0.05, 0.05]cm to [-1, 1]m). As shown in the pointwise error maps (Fig. 4C-3), JacobiNet achieves significantly higher accuracy than the baseline, with a relative $L_2$ error of 0.044, corresponding to an approximately 100× improvement.

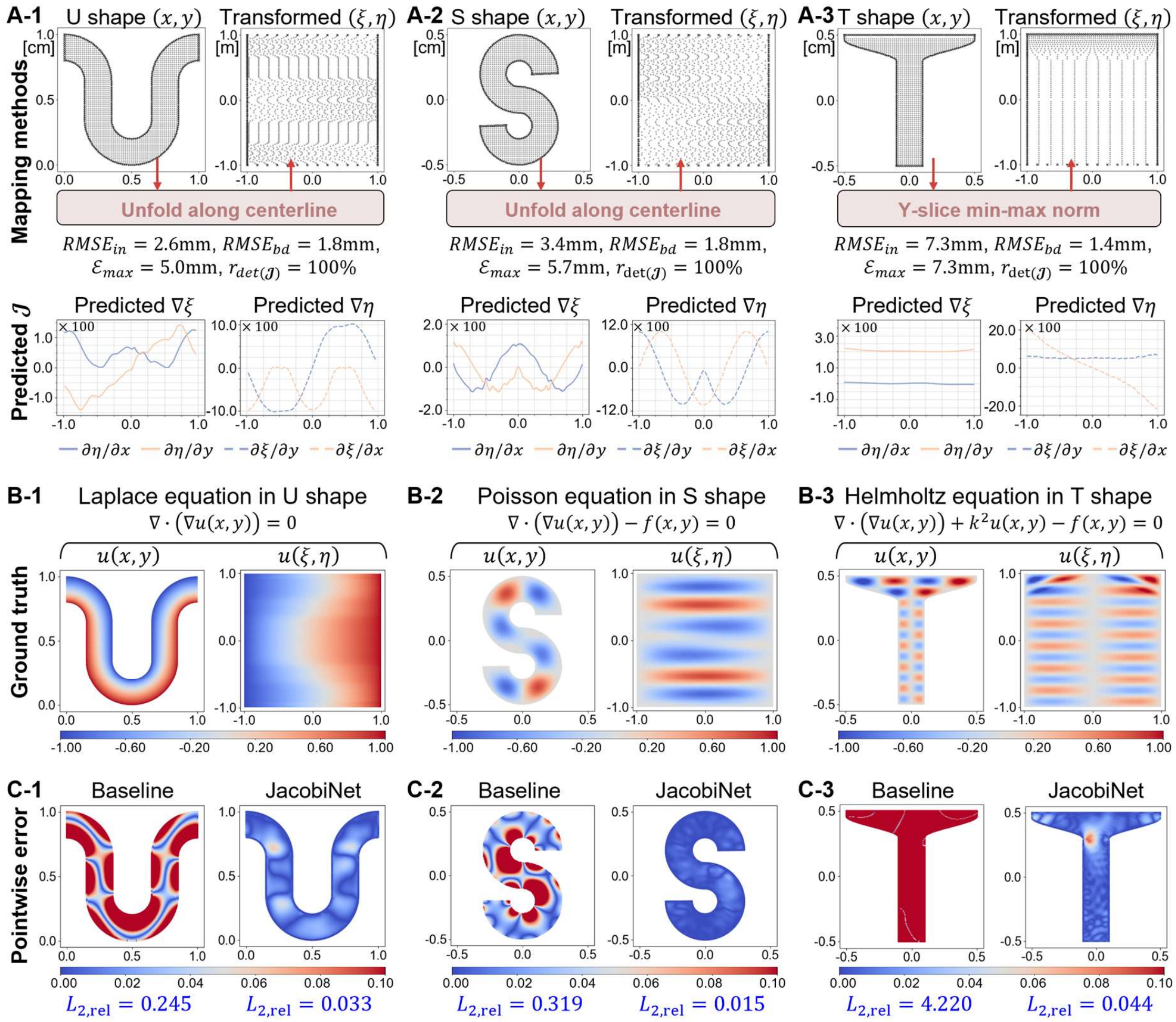

**Figure 4. JacobiNet learns diverse mappings across complex domains and enables accurate PDE solutions. (A)** Three representative geometries illustrating JacobiNet's ability to learn different geometric editing operations and map physical domains to a unit reference domain. **(B)** Ground-truth solutions of the governing equations, along with the corresponding predictions in both the physical and reference domains. **(C)** JacobiNet enables accurate PDE solutions in complex geometries and improves prediction accuracy compared with the baseline method.

### 3.4 Navier-Stokes equation in 3D stenosed vessels

Vascular stenosis, primarily caused by atherosclerotic plaque buildup, leads to abnormal narrowing of blood vessels and is a major contributor to cardiovascular diseases [45, 46]. The blood flow within a stenosed vessel can be described by the incompressible Navier–Stokes (NS) equations, which govern the conservation of mass and momentum in fluid dynamics. The incompressible NS equations, expressed in their steady and non-dimensional form, are given by

$$\nabla \cdot \vec{u} = 0, \tag{26}$$

$$\vec{u} \cdot \nabla \vec{u} + \nabla p - \frac{1}{Re} \nabla^2 \vec{u} = 0, \tag{27}$$

where $\rho$ and $\mu$ are the density and viscosity of blood flow [47], with values of $1060\ \mathrm{kg/m^3}$ and $0.0035\ \mathrm{Pa \cdot s}$, respectively. The boundary conditions include a velocity inlet, a pressure outlet, and no-slip conditions on vessel walls, detailed as follows:

$$U = U_{max}\left(1 - \frac{r^2}{R^2}\right), \text{ on } \Gamma_{inlet}; \tag{28}$$

$$p = p_{outlet}, \text{ on } \Gamma_{outlet}; \tag{29}$$

$$U = 0\ , \text{ on } \Gamma_{wall}, \tag{30}$$

here, $U_{max}$ is set to 0.50m/s, imposed along the normal direction of the inlet cross-section, corresponding to a Reynolds number of Re=350, which is representative of common physiological blood flow conditions [47]. The vessel radius at the inlet is defined as $R$, and $r$ denotes the radial distance from the vessel centerline, given by $r = \sqrt{x^2 + y^2}$. The outlet pressure $p_{outlet}$ is set to zero.

The stenosis is modelled within a standard left main coronary artery (LM) geometry [48], with a diameter of $D = 4.69\ mm$ and a length of $L = 10.00\ mm$. Two configurations are considered: (i) concentric stenosis, characterized by symmetric narrowing around the vessel axis, and (ii) eccentric stenosis, where the narrowing is asymmetric and shifted toward one side of the vessel wall. A 50% diameter reduction is imposed, corresponding to common moderate stenosis according to the CAD-RADS reporting system [49], resulting in a minimum diameter of $d = 2.35\ mm$. The stenotic segment extends over a length of $l = 6.00\ mm$, approximately 1.3 times the vessel diameter, with the axial lumen radius prescribed by a raised-cosine ("cosine-bell") profile, consistent with previous reports [47].

In this case, the stenosed vessel geometries $(x, y, z)$ were transformed into a cylindrical domain $(\xi, \eta, \zeta)$ through geometric editing, where axial slices were extracted and each slice was radially normalized to reconstruct a uniform cylindrical shape (Fig. 5A). The ground truth was obtained from computational fluid dynamics (CFD) simulations performed in ANSYS Fluent. The computational mesh was generated in ANSYS Meshing, consisting mainly of tetrahedral elements. According to mesh-independence testing (Fig. 5B), the element size was fixed at 0.05 mm. A prism-based boundary-layer mesh with 10 inflation layers was generated near the vessel wall to resolve near-wall flow gradients. The first layer thickness was set to 0.01mm, and the layer thickness increased with a growth rate of 1.1. The resulting mesh contained 2,613,357 elements for the concentric stenosis model and 2,622,278 elements for the eccentric stenosis model. Numerical simulations employed the pressure-based solver with a laminar viscous model, consistent with a Reynolds number of 350. Pressure–velocity coupling was treated with the coupled scheme, while spatial discretization used the least squares cell-based gradient evaluation. The pressure field was discretized with a second-order scheme, and the momentum equations were solved using the second-order upwind method. Solutions were advanced to convergence with residuals $< 10^{-6}$ and stable monitored quantities (e.g., wall shear stress). The resulting ground truth results are shown in Fig. 5C.

On this transformed domain, hard-constrained boundary conditions were imposed, and the trial functions were constructed as follows:

$$\hat{u}(x, y, z) = \phi_{wall}(1 - \phi_{in})\mathcal{N}_u^L(x, y, z), \tag{31}$$

$$\hat{v}(x, y, z) = \phi_{wall}(1 - \phi_{in})\mathcal{N}_v^L(x, y, z), \tag{32}$$

$$\widehat{w}(x, y, z) = \phi_{wall}(1 - \phi_{in})\mathcal{N}_w^L(x, y, z) + \phi_{wall}\phi_{in}U_{max}, \tag{33}$$

$$\hat{p}(x, y, z) = \phi_{in}\mathcal{N}_p^L(x, y, z), \tag{34}$$

where $\hat{u}$, $\hat{v}$, $\widehat{w}$, $\hat{p}$ denote the final predictions for the velocity and pressure fields, and $\mathcal{N}_u^L$, $\mathcal{N}_v^L$, $\mathcal{N}_w^L$, $\mathcal{N}_p^L$ represent the neural network outputs. Here, $\phi_{wall}$ enforces the no-slip condition at the vessel walls, while $\phi_{in}$ imposes the inlet and outlet constraints, both defined in the transformed domain as:

$$\phi_{wall} = 1 - (\xi^2 + \eta^2), \tag{35}$$

$$\phi_{in} = \frac{1-\zeta}{2}, \tag{36}$$

On the vessel wall, where $\xi^2 + \eta^2 = 1$, the $\phi_{wall}$ is thus uniformly zero, leading to predicted velocities $\hat{u}$, $\hat{v}$, $\widehat{w}$=0, thereby satisfying the no-slip boundary condition. At the inlet and outlet, corresponding to $\zeta = -1$ and $\zeta = 1$, the $\phi_{in}$ are then 1 and 0, respectively, which enforce the parabolic inlet profile $\widehat{w} = U_{max}\,(1 - (\xi^2 + \eta^2))$ and the zero-pressure outlet condition.

As shown in Fig. 5C, JacobiNet accurately reproduces the FVM velocity and pressure fields, achieving markedly lower pointwise errors than the baseline PINN, especially near the stenosis throat. Table 1 further quantifies this improvement: the overall $L_{2,\text{rel}}$ error is reduced from 0.11–0.47 (baseline) to 0.01–0.09 (JacobiNet) across the full domain, and from 0.09–0.88 to 0.01–0.16 on representative slices. These results highlight JacobiNet's superior accuracy and robustness in modeling blood flow in complex vascular geometries compared to baseline PINNs.

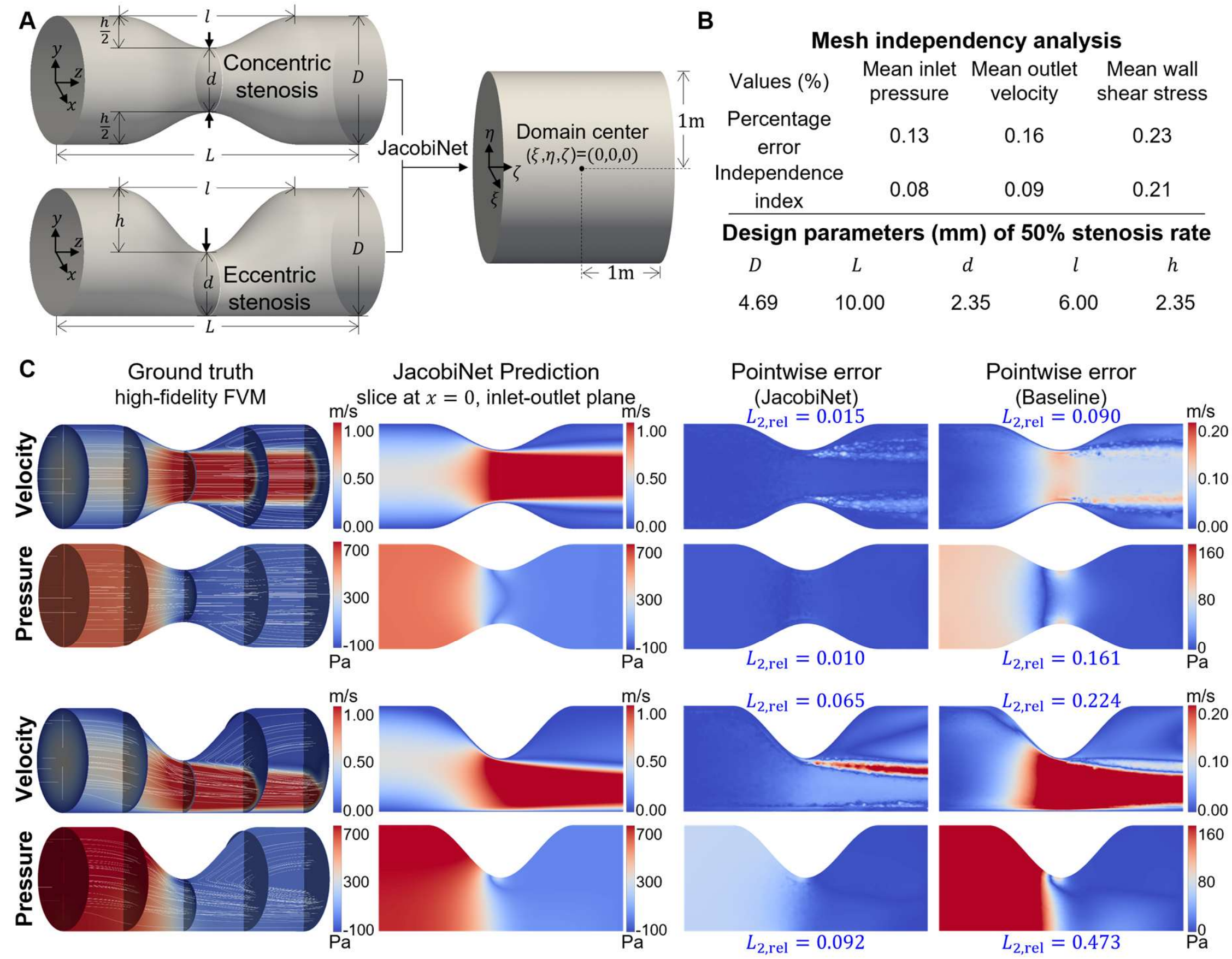

**Figure 5. JacobiNet performance in 3D stenosed vessels. (A)** Concentric and eccentric stenosed geometries with design parameters. **(B)** Mesh-independency analysis and stenosis configuration. **(C)** Ground truth FVM solutions, JacobiNet predictions (slice at $x = 0$, inlet–outlet plane), and corresponding pointwise errors (JacobiNet vs. Baseline).

**Table 1. Relative $L_2$ errors (velocity / pressure) on the 3D stenosis vessels (JacobiNet vs. Baseline)**

| | All domain | | Slice at $z = L/2$ (minimum lumen area) | | Slice at $x = 0$ (inlet–outlet plane) | |
|---|---|---|---|---|---|---|
| Stenosis type | Baseline | JacobiNet | Baseline | JacobiNet | Baseline | JacobiNet |
| Concentric stenosis | 0.11/0.16 | 0.02/0.01 | 0.12/0.88 | 0.01/0.13 | 0.09/0.16 | 0.02/0.01 |
| Eccentric stenosis | 0.31/0.47 | 0.09/0.09 | 0.28/0.46 | 0.04/0.16 | 0.22/0.47 | 0.07/0.09 |

## 3.5 Navier-Stokes equation in vessel-like shapes with varying lengths and deformations

In the previous cases, we adopted a case-specific training scheme, where each geometry was paired with its own individually trained JacobiNet. While this approach is efficient—requiring less than 67s per case, as shown in Fig. 2—it does not fully leverage the neural network's generalization capability. For practical applications such as medical imaging and real-time surgical planning, generalization across a family of structurally related geometries is essential.

To evaluate JacobiNet's geometric generalization and inference capability, we construct a dataset of synthetic vessel-like domains that mimic common cardiovascular pathologies, including stenosis and aneurysm. We generate 250 vessel-like geometries (200 stenoses, 50 aneurysms) using a template based on the diameter and length of the left coronary main artery [48], with perturbations spanning a wide range of lengths (4–17 mm) and deformation severity (−90% to +30%). The dataset distribution, shown in Fig. 6B, aligns with prior studies on coronary morphologies [48]. Samples are split into 70% training, 15% validation, and 15% testing. After training on the training/validation sets, JacobiNet is evaluated on unseen test geometries to perform domain mapping and solve the steady Navier–Stokes equations.

It is important to note that, unlike the case-specific training scheme used previously, the generalization setup requires JacobiNet to share network parameters across multiple geometries. This introduces a key problem as directly inputting only the $(x, y)$ coordinates can result in ambiguous mappings, where the same spatial point corresponds to different $(\xi, \eta)$ values across different geometries. To address this, we incorporate a simple feature extraction module to automatically encode geometric characteristics—such as vessel length and deformation severity—into a feature vector for each case. This vector, serving as an identifier for each case, is then concatenated with the spatial coordinate $(x, y)$ and jointly fed into JacobiNet. As illustrated in Fig. 6A and Fig. 6C, this input augmentation enables the network to learn context-aware transformations and effectively distinguish between geometric instances. Results show that JacobiNet achieves low RMSE (< 0.001 m) within short training times—6233.5 seconds for the stenosis group and 1298.9 seconds for the aneurysm group. Once trained, JacobiNet enables real-time inference on unseen geometries within the same morphological family, with each forward pass taking only milliseconds. For two sample unseen geometries (Fig. 6E), JacobiNet predicts natural coordinates with errors as low as 4E−4m in 34.07ms and 7E-4m in 44.27ms, respectively. These results demonstrate the model's strong geometric generalization and efficient learning capability for shape-varying domains.

For two-dimensional blood flow, the incompressible steady Navier–Stokes equations and boundary conditions as in Eqs. (26)-(30) are adopted in 2D form. The inlet peak velocity $U_{max}$ is varied from 0.0106m/s to 1.06m/s corresponding to Reynolds numbers *Re* = 10 to 1000, which are representative of physiological blood flow conditions.

For two sample unseen geometries (51.2% narrowing and 28.4% expansion), JacobiNet is evaluated under five Reynolds numbers (*Re* = 10, 100, 300, 500, 1000). The predicted velocity magnitudes, pressure fields, and streamlines show excellent agreement with CFD results, with relative $L_2$ errors ranging from 0.006–0.069 for velocity and 0.029–0.237 for pressure. At *Re* = 10, JacobiNet captures wall-hugging streamlines, where streamlines adhere closely to the vessel walls, while it accurately resolves near-wall reverse flow at Re = 1000.

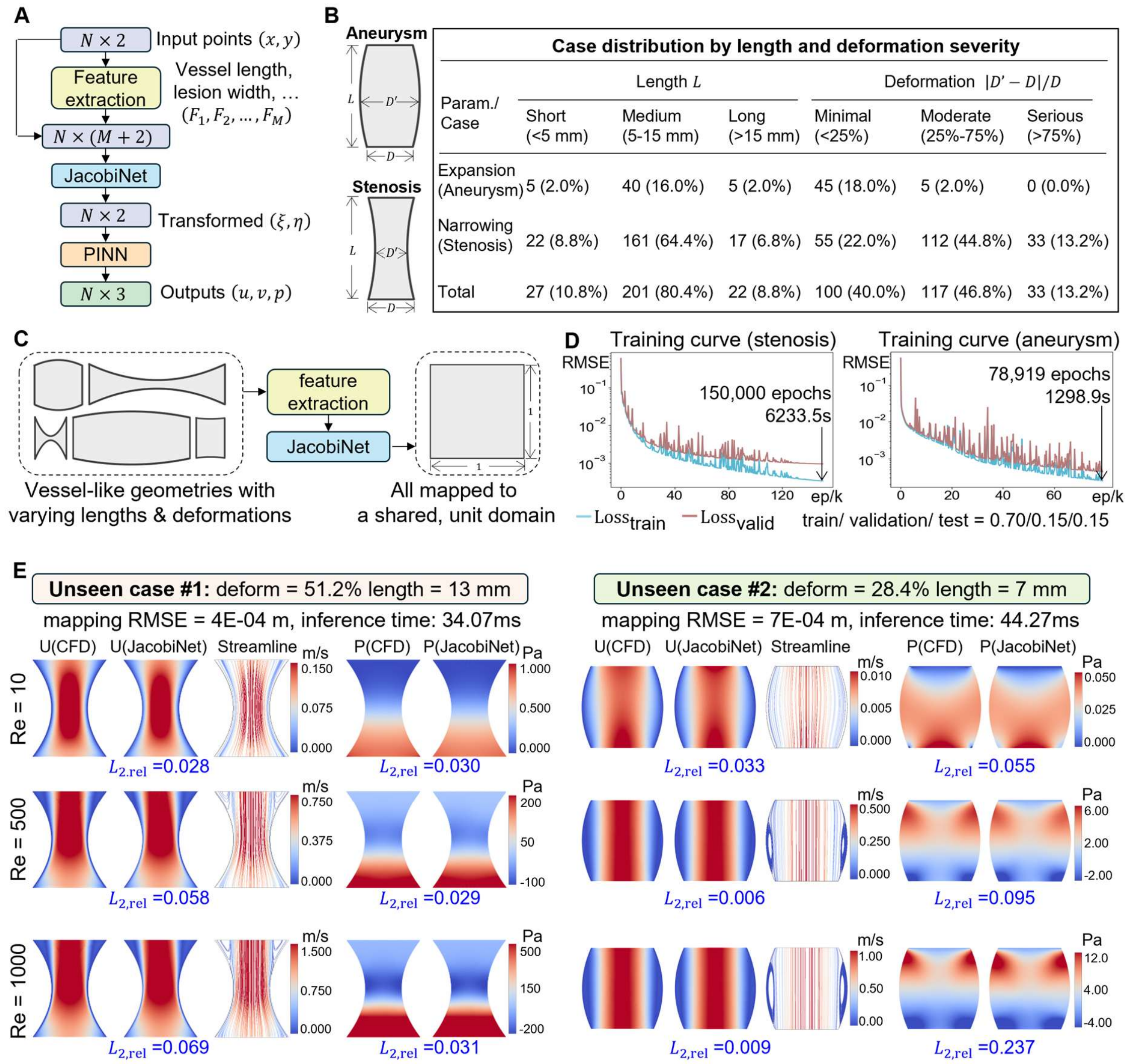

**Figure 6. JacobiNet provides a generalizable framework that maps vessel-like geometries with varying lengths and deformations to a shared, unit domain, enabling accurate flow prediction with PINN. (A)** Overview of JacobiNet + PINN framework for parametric geometries. **(B)** Case distribution of the synthetic vessel-like dataset. **(C)** Vessel cases with varying $L$ and $|D' - D|/D$ are encoded via feature extraction, enabling JacobiNet to map all geometries into a shared unit domain. **(D)** Training curves for stenosis and aneurysm cases. **(E)** Predictions of JacobiNet + PINN on unseen geometries for the Navier–Stokes equations over varying Reynolds number (Re = 10, 500, and 1000).

## 3.6 Computational overhead analysis

To further quantify the additional cost introduced by JacobiNet, we compare it with the baseline by measuring the wall-clock time of four components in each training step: the forward pass, the loss computation, the backpropagation, and the total runtime. Table 2 reports the averaged runtime per epoch (mean ± standard deviation, in milliseconds) over 100 repetitions for representative test cases.

**Table 2. Computational overhead comparison of Baseline PINNs vs. JacobiNet (ms/epoch)**

| Test case | Methods | Forward pass | Loss computation | Back-propagation | Total runtime | Percentage increase |
|---|---|---|---|---|---|---|
| | Baseline | 0.16 ± 0.02 | 5.80 ± 0.54 | 6.25 ± 2.50 | 12.21 ± 2.69 | - |
| Laplace Eq. in U shape | JacobiNet | 0.27 ± 0.01 | 6.27 ± 0.98 | 7.63 ± 0.75 | 14.17 ± 0.90 | 16.04% |
| Poisson Eq. in S shape | Baseline | 0.14 ± 0.01 | 4.07 ± 1.30 | 5.80 ± 2.39 | 10.01 ± 2.07 | - |
| | JacobiNet | 0.26 ± 0.01 | 5.15 ± 1.83 | 6.25 ± 0.43 | 11.66 ± 1.84 | 16.57% |
| Helmholtz Eq. in T shape | Baseline | 0.12 ± 0.02 | 3.00 ± 0.71 | 5.10 ± 0.22 | 8.22 ± 0.73 | - |
| | JacobiNet | 0.25 ± 0.01 | 3.73 ± 0.59 | 5.70 ± 0.45 | 9.68 ± 0.43 | 17.76% |
| 3D NS Eqs. in concentric stenosis | Baseline | 0.21 ± 0.00 | 22.73 ± 2.82 | 29.60 ± 0.89 | 52.55 ± 2.65 | - |
| | JacobiNet | 0.47 ± 0.00 | 30.42 ± 1.85 | 46.97 ± 2.82 | 77.86 ± 4.28 | 48.17% |
| 3D NS Eqs. in eccentric stenosis | Baseline | 0.29 ± 0.02 | 24.91 ± 3.61 | 37.14 ± 3.43 | 62.34 ± 3.90 | - |
| | JacobiNet | 0.48 ± 0.02 | 34.26 ± 4.52 | 59.24 ± 2.16 | 93.98 ± 6.60 | 50.75% |

Across all cases, the forward pass remains relatively inexpensive (typically <0.5ms), while the loss computation and backpropagation dominate the runtime. Although JacobiNet eliminates the need for computing boundary-loss terms through hard constraints, the enlarged computation graph and additional trial-function operations still increase the computational burden. Specifically, for scalar-field problems, this increase is relatively modest: the total runtime rises from 12.21/10.01/8.22ms to 14.17/11.66/9.68ms (+16.8% on average). In contrast, for more complex 3D stenosed-vessel cases, the overhead becomes more pronounced. Owing to the inherent complexity of the Navier–Stokes equations and the necessity of applying trial functions to each network output component—even under vectorized parallel operations—the runtime increases substantially, with concentric stenosis growing from 52.55ms to 77.86ms (+48.2%) and eccentric stenosis rising from 62.34ms to 93.98ms (+50.8%).

Nevertheless, the added cost remains manageable relative to the overall training runtime, especially considering the average accuracy gains of approximately one order of magnitude consistently achieved by JacobiNet. In the next section, we further demonstrate the efficiency improvement brought by JacobiNet: to reach the same accuracy level, JacobiNet requires much fewer training epochs, thereby yielding a substantial reduction in the total training time.

## 3.7 Comparison with state-of-the-art methods

In this section, we compare the proposed JacobiNet with a baseline PINN and several state-of-the-art approaches, including:

- a baseline PINN with standard non-dimensionalization [25] for input normalization,
- a gradient-based reweighting scheme (Gradient Normalization, Grad Norm) to mitigate loss imbalance [18],
- a random Fourier feature (RFF) embedding strategy to alleviate spectral bias [44],
- residual-based adaptive sampling (RAR-D) [33], which dynamically redistributes collocation points according to the magnitude of the PDE residuals, allowing the training process to focus on regions with larger approximation errors.

- Δ-PINN [50], which introduces Laplacian eigenfunction–based feature embeddings to enhance the representation of spatial structures and improve the learning capability of PINNs in complex domains.

Using the Navier–Stokes equations in vessel-like tube domains as benchmark cases, JacobiNet consistently demonstrates superior performance, particularly at higher Reynolds numbers (Re > 100). Compared to the baseline PINN, JacobiNet achieves 2.83× and 4.46× lower relative $L_2$ errors for velocity and pressure predictions, as shown in Table 3, 4. Even when compared with the best-performing state-of-the-art method (Δ-PINN), JacobiNet achieves further improvements of 1.37× in velocity and 1.72× in pressure accuracy on averages. Notably, unlike Δ-PINN, which relies on mesh-based finite element discretization to construct eigenfunction features [50], JacobiNet adopts a data-driven coordinate transformation framework, which learns a differentiable mapping together with its associated Jacobian directly from point cloud data. This avoids the need for mesh-based computation and enables a unified differentiable pipeline for geometry handling and PDE solving.

**Table 3. Relative $L_2$ errors on the stenosis benchmark (51.2% Deformation, 13.0 mm Length)**

Each value is reported as velocity / pressure relative $L_2$ errors. In each column, Green indicates the best performance, while orange denotes the worst result among the compared methods.

| | $L_{2.\text{rel}}$ OF TEST STENOSIS (DEFORM=51.2%, LENGTH=13.0mm) | | | | |
|---|---|---|---|---|---|
| BENCHMARK #1 | *Re = 10* | *Re = 100* | *Re = 300* | *Re = 500* | *Re = 1000* |
| Baseline non-dim | 0.034/ 0.041 | 0.037/ 0.065 | 0.225/ 0.116 | 0.248/ 0.110 | 0.210/ 0.097 |
| Grad Norm | 0.036/ 0.039 | 0.053/ 0.053 | 0.047/ 0.054 | 0.230/ 0.106 | 0.203/ 0.097 |
| RFF | 0.020/ 0.022 | 0.016/ 0.013 | 0.122/ 0.058 | 0.143/ 0.068 | 0.142/ 0.062 |
| RAR-D | 0.027/ 0.039 | 0.053/ 0.051 | 0.176/ 0.096 | 0.178/ 0.097 | 0.190/ 0.089 |
| Δ-PINN | 0.033/ 0.021 | 0.008/ 0.012 | 0.067/ 0.032 | 0.153/ 0.073 | 0.172/ 0.084 |
| JacobiNet | 0.028/ 0.032 | 0.011/ 0.016 | 0.035/ 0.025 | 0.058/ 0.029 | 0.083/ 0.031 |
| JacobiNet + Grad Norm | 0.022/ 0.024 | 0.009/ 0.017 | 0.057/ 0.034 | 0.089/ 0.043 | 0.079/ 0.025 |
| JacobiNet + RFF | 0.017/ 0.021 | 0.004/ 0.011 | 0.025/ 0.030 | 0.029/ 0.014 | 0.014/ 0.006 |

**Table 4. Relative $L_2$ errors on the aneurysm benchmark (28.4% Deformation, 7.00 mm Length)**

Each value is reported as velocity / pressure relative $L_2$ errors. In each column, Green indicates the best performance, while orange denotes the worst result among the compared methods.

| | $L_{2.\text{rel}}$ OF TEST ANEURYSM (DEFORM=28.4%, LENGTH=7.00mm) | | | | |
|---|---|---|---|---|---|
| BENCHMARK #2 | *Re = 10* | *Re = 100* | *Re = 300* | *Re = 500* | *Re = 1000* |
| Baseline non-dim | 0.024/ 0.102 | 0.005/ 0.105 | 0.026/ 1.030 | 0.020/ 1.020 | 0.019/ 1.162 |
| Grad Norm | 0.024/ 0.097 | 0.006/ 0.114 | 0.030/ 1.150 | 0.022/ 1.090 | 0.039/ 1.140 |
| RFF | 0.022/ 0.171 | 0.003/ 0.209 | 0.018/ 0.778 | 0.010/ 0.561 | 0.017/ 0.864 |
| RAR-D | 0.017/ 0.130 | 0.004/ 0.161 | 0.021/ 0.963 | 0.018/ 0.832 | 0.023/ 1.001 |
| Δ-PINN | 0.021/ 0.074 | 0.003/ 0.133 | 0.009/ 0.123 | 0.008/ 0.359 | 0.014/ 0.538 |
| JacobiNet | 0.033/ 0.055 | 0.008/ 0.184 | 0.007/ 0.114 | 0.006/ 0.095 | 0.009/ 0.237 |
| JacobiNet + Grad Norm | 0.032/ 0.059 | 0.007/ 0.060 | 0.007/ 0.097 | 0.007/ 0.089 | 0.008/ 0.193 |
| JacobiNet + RFF | 0.034/ 0.047 | 0.005/ 0.115 | 0.005/ 0.140 | 0.005/ 0.154 | 0.006/ 0.132 |

As a lightweight front-end, JacobiNet can be effortlessly integrated with state-of-the-art frameworks in a plug-and-play manner, consistently enhancing accuracy and efficiency. At $Re$ = 1000, the JacobiNet+RFF model achieves the best velocity prediction accuracy among all tested methods. Compared to the RFF-only model, the hybrid model reduces velocity and pressure errors by 10.1× and 10.3× in the stenosis case, and by 2.83× and 6.55× in the aneurysm case, respectively. In terms of efficiency, at $Re$ = 500, to reach the same accuracy that the RFF-only model achieves in 20,000 epochs, JacobiNet+RFF requires only 510 iterations for stenosis and 3370 iterations for aneurysm, representing a 39.2× and 5.93× speed-up, respectively. A detailed comparison is provided in Fig. 7.

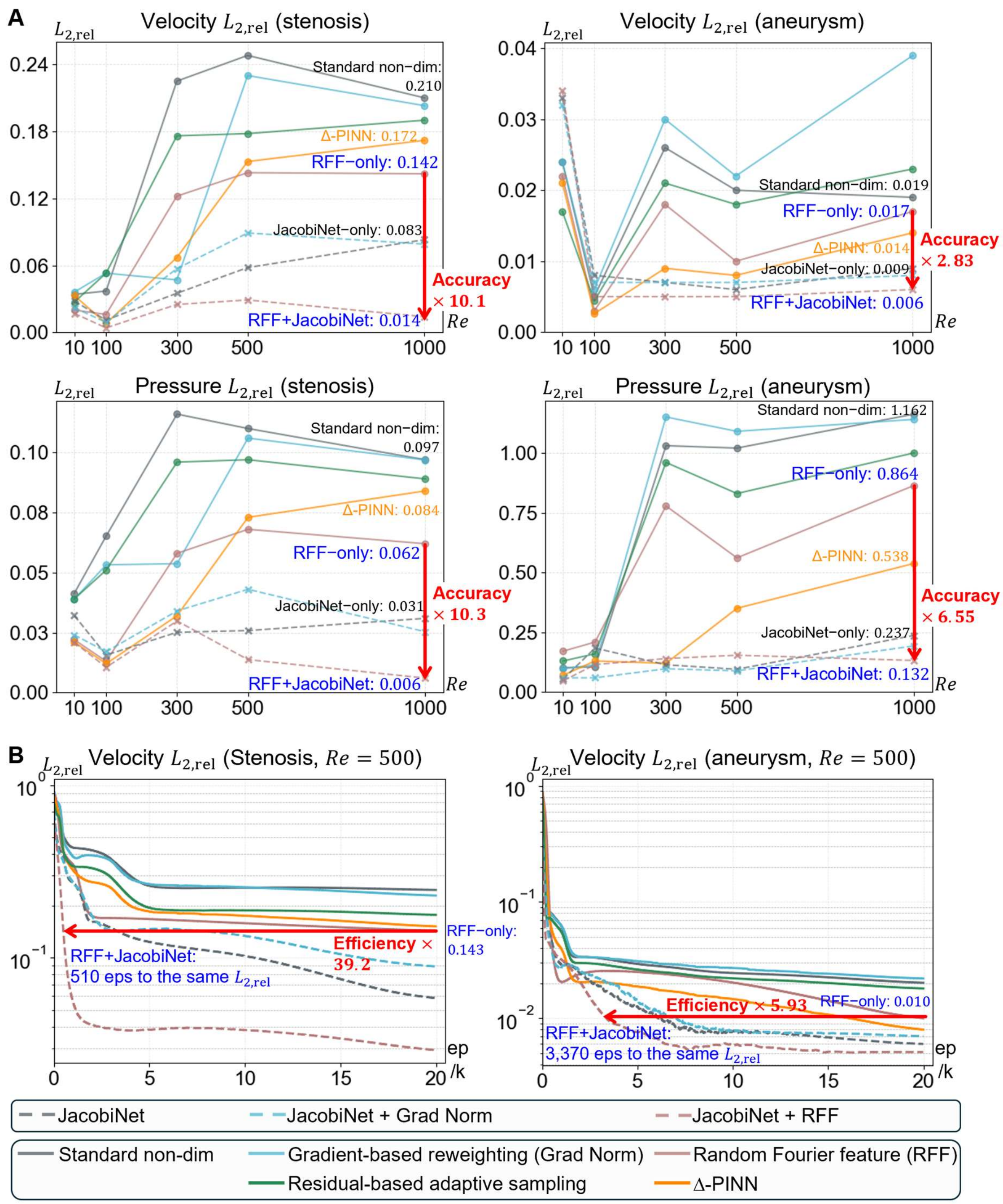

**Figure 7. Quantitative comparison of JacobiNet and baseline & state-of-the-art methods. (A)** Improvement in accuracy. **(B)** Improvement in efficiency.

# 4. Discussion & Conclusion

In this work, we started from the observation that PINNs often suffer from instability and slow convergence when applied to complex domains. We traced these challenges to three fundamental issues, namely inconsistent normalization, inaccurate boundary enforcement, and imbalanced loss-term competition. We further highlighted the limitations of existing coordinate-transformed workflows, where mappings are either case-specific, mesh-based and non-differentiable (numerical methods) or decoupled from the physics solver (learning-based parameterizations). To address these issues, we proposed JacobiNet, an end-to-end differentiable framework that integrates domain mapping and PDE solving within a single computational graph. JacobiNet automatically handles Jacobians and residuals without requiring PDE reformulation via chain rules, while enabling hard boundary enforcement through trial functions defined in the mapped domain. Extensive evaluations—including ablation studies, computational overhead analysis, and comparison with state-of-the-art methods—demonstrated that JacobiNet achieved notable improvements in both accuracy and efficiency, with average improvements exceeding an order of magnitude. In summary, JacobiNet learns continuous, differentiable mappings that unify geometric preprocessing with physical modeling, enabling scalable coordinate transformations for PINN-based scientific computing.

Although JacobiNet improves the efficiency of PINNs in complex domains and enables real-time geometric mapping, the subsequent PDE solving still relies on the iterative optimization inherent to PINNs, leaving the overall inference speed constrained by the downstream physical network. Besides, our computational overhead analysis further shows that a shallow 2×128 JacobiNet introduces over 20% additional cost in subsequent PINN computations, rising to nearly 50% for three-dimensional Navier–Stokes problems—particularly in loss computation and backpropagation. Nevertheless, this overhead remains manageable relative to the total runtime, especially given the substantial accuracy gains achieved. As demonstrated in Section 3.7, JacobiNet reduces the number of training epochs required to reach a given accuracy by up to an order of magnitude, yielding a net improvement in overall training efficiency despite the per-epoch cost. Beyond fully connected MLPs, mapping irregular domains to regular reference spaces enables the adoption of efficient classical architectures such as CNNs, whose locality, weight sharing, and parallelism offer promising avenues for further acceleration [28]. The value of JacobiNet also extends well beyond the PINN framework: neural operator methods that require geometric consistency and coordinate normalization—such as Fourier Neural Operator (FNO) [30, 51], DeepONet [52, 53], PINO [54], or even graph neural networks [55] with positional encoding—can adopt JacobiNet as a preprocessing normalization module to construct a well-structured, differentiable, and transferable input space.

In contrast to the exact imposition of Dirichlet boundary conditions via trial functions, which has been explored in the literature [27], hard enforcement of Neumann boundary conditions remains challenging. Incorporating Neumann conditions into the trial function would require PINNs to predict not only the solution but also its derivatives, thereby introducing gradient-consistency losses and increasing model complexity. Treating them in a soft manner therefore provides a practical balance between accuracy and efficiency. In future work, we aim to extend this framework to incorporate hard-constrained Neumann boundary conditions, potentially leveraging auxiliary gradient networks [20, 56], and systematically investigate the associated trade-offs.

Besides, it is important to note that loss-term imbalance in PINNs is not solely determined by geometric complexity, but also depends on the underlying physics of the governing equations. While the proposed framework alleviates this issue by improving geometric conditioning and removing boundary-related loss terms via reference-domain trial functions, it does not fully eliminate loss imbalance in general PINN training. For example, in 3D Navier–Stokes problems, the governing system consists of one continuity equation and three momentum equations with strong nonlinear coupling. The interaction among these coupled PDE residuals can still lead to significant loss-term competition during training. This issue may become more pronounced at higher Reynolds numbers, where increasingly complex flow features and sharper local velocity gradients make the flow fields more difficult to predict accurately, as reflected by the increased prediction errors of several baseline and comparison methods in Fig. 7A. These observations emphasize that improving the robustness of JacobiNet-PINN for high-Reynolds-number flows will be an important direction for our future work.

In domains with higher geometric complexity—such as 3D vascular trees, multiconnected structures, or topologically discontinuous shapes—the construction of reliable supervised mapping data becomes more challenging, and the downstream PINN solver may also be more susceptible to failure. While the geometric editing operations explored in this work are single step and supervised by training pairs, extending JacobiNet with multi-step pipeline or unsupervised learning strategies [38, 41] may further broaden its generalization capacity and practical applicability. Recent advances in integral volumetric parameterization via deep neural networks [36, 39] provide powerful strategies for handling highly complex geometries. Combining these methods with our proposed framework could further strengthen the development of PINN-based modeling in complex domains. Importantly, although we adopted a simple 2×128 MLP as the geometric processing module in this study, this component could in principle be replaced by more advanced architectures [39, 41] tailored for mapping or representation learning, underscoring the flexibility and extensibility of JacobiNet.

## Conflict of Interest

The authors declare that the research was conducted in the absence of any commercial or financial relationships that could be construed as a potential conflict of interest.

## Data availability

Codes and datasets used in this study will be made publicly available upon publication at https://github.com/xchenim/JacobiNet.

## Acknowledgements

This work was supported by the HKUST Startup Funding, Guangdong Natural Science Foundation GDST25EG07.

## Contributions

X. C., T. Z., Z. R., W. H. proposed and designed the research. X. C., J. Y. carried out theoretical analysis and mathematical modeling. Numerical experiments, data processing and visualization were conducted by X. C., J. Y., J. Z. All authors participated in discussions, wrote the manuscript, and approved the final version.

# Appendix-1

## Experimental Validation of Loss-term Imbalance in Complex Geometries

To quantitatively verify the loss-term imbalance in complex domains, we designed a controlled numerical experiment. The test set consists of a reference unit square domain $[-1\text{m}, 1\text{m}]^2$ and a series of domains with progressively increasing geometric irregularity. The irregular domains are generated by fixing four corner points $(-1,-1)$m, $(-1,1)$m, $(1,1)$m, $(1,-1)$m, and constructing the four edges via sinusoidal perturbations applied to straight segments. Formally, the boundary curve $\gamma(s)$ for $s \in [0,4)$ is parameterized as follows:

$$\gamma(s) = \mathbf{P}_{k,:} + u\mathbf{D}_{k,:} + A\sin(\omega\pi u)\,\mathbf{N}_{k,:}, \tag{37}$$

where $k = \lfloor s \rfloor \in \{0,1,2,3\}$ indexes the bottom, right, top, and left edges, and $u = s - k \in [0,1)$ is the local parameter along edge $k$. Here, $\mathbf{P}_{k,:}$, $\mathbf{D}_{k,:}$, $\mathbf{N}_{k,:}$ denote the starting point, tangent direction, and the normal direction of the unperturbed edge, respectively, with the normal direction used for applying the sinusoidal perturbation. The matrices **P**, **D**, and **N** are given by:

$$\mathbf{P} = \begin{bmatrix} -1 & -1 \\ 1 & -1 \\ 1 & 1 \\ -1 & 1 \end{bmatrix}, \mathbf{D} = \begin{bmatrix} 2 & 0 \\ 0 & 2 \\ -2 & 0 \\ 0 & -2 \end{bmatrix}, \mathbf{N} = \begin{bmatrix} 0 & -1 \\ 1 & 0 \\ 0 & 1 \\ -1 & 0 \end{bmatrix}. \tag{38}$$

Each case adopts different combinations of amplitudes $A$ and frequencies $\omega$ to represent different levels of boundary irregularity. When $A = 0$ and $\omega = 0$, the sinusoidal perturbation vanishes and the geometry degenerates to the standard unit square domain $[-1\text{m}, 1\text{m}]^2$.

For all cases, points are uniformly sampled with an interval of 0.05m. The amplitude $A$, frequency $\omega$, domain shape, and the number of sampled points (boundary and internal) for each case are summarized in Fig. 8A. The baseline PDE is the Poisson equation, with boundary conditions specified in Eqs. (16)-(18) of the main text. The ground truth solutions are obtained using the finite volume method on high-fidelity meshes, as illustrated in Fig. 8B.

The PINN is implemented as a fully connected MLP with three hidden layers, each comprising 64 neurons, and employs the SiLU activation function. The network is trained using the Adam optimizer with an initial learning rate of $10^{-3}$, combined with a cosine annealing scheduler that gradually reduces the learning rate to $10^{-5}$ over the course of training. All experiments are conducted for 3000 epochs.

During training, the following quantitative metrics are monitored to assess the severity of the loss-term imbalance and its impact on optimization:

1. Loss ratio:

$$\mathcal{R}_{loss}(t) = \frac{\mathcal{L}_u(t)}{\mathcal{L}_b(t)}, \tag{39}$$

where $\mathcal{R}_{loss}(t)$ is the ratio between the PDE residual loss $\mathcal{L}_u$ and the boundary condition loss $\mathcal{L}_b$. Here, $t$ denotes the training iteration. The loss-ratio curves characterize the difference in numerical scale between the two loss terms.

2. Loss gradient-norm ratio:

$$\mathcal{R}_{grad}(t) = \frac{\|\nabla_\vartheta \mathcal{L}_u(t)\|_2}{\|\nabla_\vartheta \mathcal{L}_b(t)\|_2}, \tag{40}$$

where $\mathcal{R}_{grad}(t)$ computes the ratio of the gradient norms of the loss terms with respect to the trainable parameters $\vartheta$. This metric reflects the relative influence of each loss term on parameter updates.

3. Relative $L_2$ error

$$L_{2,\mathrm{rel}} = \frac{\sqrt{\sum_{i=1}^{N}[u(x_i,y_i)-\hat{u}(x_i,y_i)]^2}}{\sqrt{\sum_{i=1}^{N}[u(x_i,y_i)]^2}}, \tag{41}$$

The final accuracy is measured by the relative $L_2$ error between the predicted solution $\hat{u}$ and the ground truth solution $u$.

The performance of PINNs across domains is shown in Figure 8C. In the unit domain, the $L_{2,\mathrm{rel}}$ error is 0.096; as the boundary complexity $(A, \omega)$ increases, the error steadily grows, reaching 0.947 in the complex domain with $A = 0.25$, $\omega = 11$. Figure 8D presents the loss ratio and loss gradient-norm ratio curves during training. In the unit domains, the different loss terms converge rapidly and remain close to 1 in gradient contribution. In contrast, in complex domains, the numerical scale of the PDE loss and boundary loss diverges substantially, with $\mathcal{R}_{loss}(t)$ reaching 0.02 at 3000 epochs—meaning that the BC term dominates the total loss, accounting for 98.06% of its magnitude. Moreover, the loss gradient-norm ratio in complex domains exhibits pronounced and irregular oscillations, indicating unstable relative contributions from the PDE residuals and boundary conditions. This alternating dominance leads to conflicting gradients, impairs convergence, and directly contributes to the error amplification observed in Figure 8C, ultimately resulting in elevated $L_{2,\mathrm{rel}}$ errors.

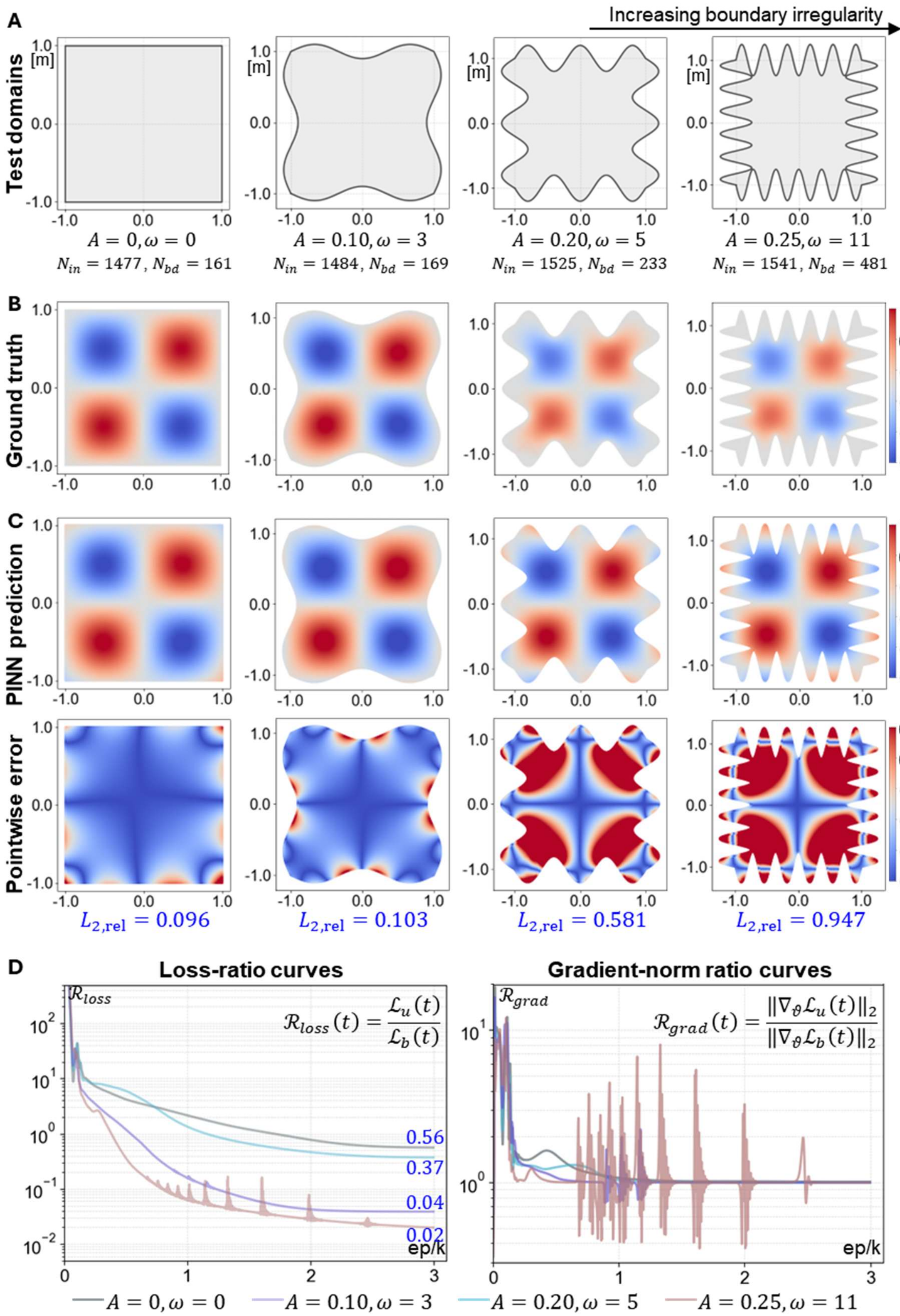

**Figure 8. Experimental validation of loss-term imbalance in complex geometries.** **(A)** Domains of increasing boundary complexity, parameterized by amplitude $A$, and frequency $\omega$. **(B)** Ground-truth PDE solutions. **(C)** PINN predictions and corresponding pointwise error fields. **(D)** Training-stage diagnostics: loss-ratio curves $\mathcal{R}_{loss}(t)$ and loss gradient-norm ratio curves $\mathcal{R}_{grad}(t)$.

## Appendix-2

### Coordinate-Transformed PINN Loss Function Reformulated via the Chain Rule

To improve the numerical stability and accuracy of PINNs when solving PDEs in complex domains, a coordinate transformation strategy is employed, which maps the irregular physical coordinates $\mathbf{x}$ onto a regular reference domain $\boldsymbol{\xi}$, as shown in Fig. 1B. Let the physical domain be denoted as $\Omega_{\mathbf{x}} \subset \mathbb{R}^D$ , and the reference domain as $\Omega_{\boldsymbol{\xi}} = [-1, 1]^D$. We therefore define a continuous bijective mapping:

$$\Phi: \mathbf{x} \mapsto \boldsymbol{\xi}(\mathbf{x}), \text{ for } \mathbf{x} \in \Omega_{\mathbf{x}}, \tag{42}$$

which establishes a one-to-one correspondence between the physical and reference domains. The Jacobian matrix of this transformation is:

$$\boldsymbol{\mathcal{J}} = \frac{\partial \boldsymbol{\xi}}{\partial \mathbf{x}} = \begin{bmatrix} \frac{\partial \xi_1}{\partial x_1} & \cdots & \frac{\partial \xi_1}{\partial x_D} \\ \vdots & \ddots & \vdots \\ \frac{\partial \xi_D}{\partial x_1} & \cdots & \frac{\partial \xi_D}{\partial x_D} \end{bmatrix}. \tag{43}$$

Now, consider a general form of a PDE defined over the physical domain $\Omega_{\mathbf{x}}$:

$$\mathcal{F}(\mathbf{x}, u, \nabla_{\mathbf{x}} u, \nabla_{\mathbf{x}}^2 u) = 0, \tag{44}$$

where $u(\mathbf{x})$ denotes the unknown scalar or vector field, $\nabla_{\mathbf{x}} u$ represents first-order spatial derivatives, and $\nabla_{\mathbf{x}}^2 u$ denotes second-order differential operators such as the Laplacian.

Through the coordinate transformation, derivatives in the physical domain can be expressed in terms of derivatives in the reference domain $\boldsymbol{\xi}$ via the chain rule. Specifically, the first-order derivatives and second-order derivatives are given by:

$$\nabla_{\mathbf{x}} u = \boldsymbol{\mathcal{J}}^{\mathrm{T}} \nabla_{\boldsymbol{\xi}} u, \tag{45}$$

$$\nabla_{\mathbf{x}}^2 \mathrm{u} = \boldsymbol{\mathcal{J}}^{\mathrm{T}} \nabla_{\boldsymbol{\xi}}^2 u \boldsymbol{\mathcal{J}} + \nabla_{\boldsymbol{\xi}} u \cdot \nabla_{\mathbf{x}}^2 \boldsymbol{\Phi}, \tag{46}$$

where $\nabla_{\mathbf{x}}^2 \boldsymbol{\Phi}$ denotes the Hessian matrices of the transformation mapping components $\boldsymbol{\xi}(\mathbf{x})$, and the dot $\cdot$ denotes tensor contraction over shared indices. Note that second-order derivatives in the physical domain are determined by the reference-domain gradients, as well as the Jacobian $\boldsymbol{\mathcal{J}}$ and its spatial derivatives. Specifically, each component of $\nabla_{\mathbf{x}}^2 \boldsymbol{\Phi}$ can be obtained by differentiating the entries of $\boldsymbol{\mathcal{J}}$ with respect to $\boldsymbol{\xi}$:

$$\nabla_{\mathbf{x}}^2 \boldsymbol{\Phi} = (\nabla_{\boldsymbol{\xi}} \boldsymbol{\mathcal{J}}) \boldsymbol{\mathcal{J}}, \tag{47}$$

as a result, the original PDE and boundary conditions in the physical domain are equivalently reformulated in the reference domain as:

$$\mathcal{F}'\left(\boldsymbol{\xi}, u, \nabla_{\boldsymbol{\xi}} u, \nabla_{\boldsymbol{\xi}}^2 u, \boldsymbol{\mathcal{J}}\right) = 0, \ \boldsymbol{\xi} \in \Omega_{\boldsymbol{\xi}}, \tag{48}$$

$$\mathcal{B}'\left(\boldsymbol{\xi}, u, \nabla_{\boldsymbol{\xi}} u, \nabla_{\boldsymbol{\xi}}^2 u, \boldsymbol{\mathcal{J}}\right) = 0, \ \boldsymbol{\xi} \in \partial\Omega_{\boldsymbol{\xi}}, \tag{49}$$

This transformed formulation allows the neural network to solve the PDE in a regular reference domain, while still accounting for the geometric complexity of the original physical space through the embedded Jacobian and its derivatives. This lays the foundation for coordinate-transformed PINN frameworks, where the neural network takes $(\xi, \eta)$ as input, and the loss terms are thus reformulated as follows:

$$\mathcal{L}_u = \sum_{j=1}^{M_u} \lambda_{u_j} \Big(\frac{1}{N_{u_j}} \sum_{i=1}^{N_{u_j}} \left\| \mathcal{F}_j' \left[ \hat{u}\left(\boldsymbol{\xi}_i^{u_j}\right), \nabla_{\boldsymbol{\xi}} \hat{u}, \nabla_{\boldsymbol{\xi}}^2 \hat{u}, \boldsymbol{\mathcal{J}_i} \right] \right\|^2 \Big), \tag{50}$$

$$\mathcal{L}_b = \sum_{k=1}^{M_b} \lambda_{b_k} \Big(\frac{1}{N_{b_k}} \sum_{i=1}^{N_{b_k}} \left\| \mathcal{B}_k' \left[ \hat{u}\left(\boldsymbol{\xi}_i^{b_k}\right), \nabla_{\boldsymbol{\xi}} \hat{u}, \nabla_{\boldsymbol{\xi}}^2 \hat{u}, \boldsymbol{\mathcal{J}_i} \right] \right\|^2 \Big), \tag{51}$$

all spatial derivatives in physical space are obtained via the chain rule (i.e., $\nabla_{\boldsymbol{\xi}} \hat{u}$, $\nabla_{\boldsymbol{\xi}}^2 \hat{u}$). The operators $\mathcal{F}_j'[\cdot]$ and $\mathcal{B}_k'[\cdot]$ are now applied on reference-domain predictions, transformed back into physical space using the Jacobian $\boldsymbol{\mathcal{J}}_i$, the Jacobian matrix of the mapping evaluated at the i-th point $\mathbf{x}_i$.

# Appendix-3

### Sensitivity Analysis of Boundary Weight $\lambda$

Accurate boundary alignment is critical for the proper enforcement of hard-constrained boundary conditions. To achieve this, we introduced a weighted boundary supervision strategy that increases the emphasis on boundary points during training, as defined in Eq. (10). This strategy inherently involves a trade-off: when the boundary weight $\lambda$ is too small, the boundary cannot be accurately captured; conversely, an excessively large $\lambda$ ensures precise boundary fitting but deteriorates the learning of interior points. In such cases, incorrectly mapped interior points may fall onto or outside the boundary, causing the trial function to be incorrectly activated in non-boundary regions, thereby destabilizing training and hindering the learning of the correct mapping.

To systematically evaluate the sensitivity of this weighting scheme, we conducted a parameter study by varying $\lambda$ from 1 (no weighting) to 100, with numerical experiments across the benchmark cases in Sections 3.1–3.4. For each choice of $\lambda$, we monitor the root mean square error $RMSE_{in}$, $RMSE_{bd}$ and the maximum normal deviation $\mathcal{E}_{max}$ of boundary points, as well as the final PDE prediction error quantified by the relative $L_2$ error, defined as Eqs. (10)-(14) of the main text.

The results are summarized in Fig. 9. The mapping error of interior points $RMSE_{in}$ increases consistently with larger $\lambda$, whereas the boundary errors ($RMSE_{bd}$, $\mathcal{E}_{max}$) decrease but gradually saturate as $\lambda$ increases. This trade-off is also evident in the overall PDE prediction accuracy: both boundary errors and the relative $L_2$ error of the PDE solution exhibit a U-shaped trend, improving at moderate values of $\lambda$ but deteriorating when $\lambda > 20$. As a result, the relative $L_2$ error reaches its lowest and most stable values within the range $\lambda \in [5,20]$. Based on this analysis, we set $\lambda = 10$ in our study, which provides a favourable balance between accurate boundary alignment and robust interior mapping. Across all reported experiments, $RMSE_{bd}$ remained below $2 \times 10^{-3}$m and $\mathcal{E}_{max}$ below $8 \times 10^{-3}$m, corresponding to 0.10% and 0.40% of the reference domain size, respectively, thereby demonstrating consistently accurate boundary alignment.

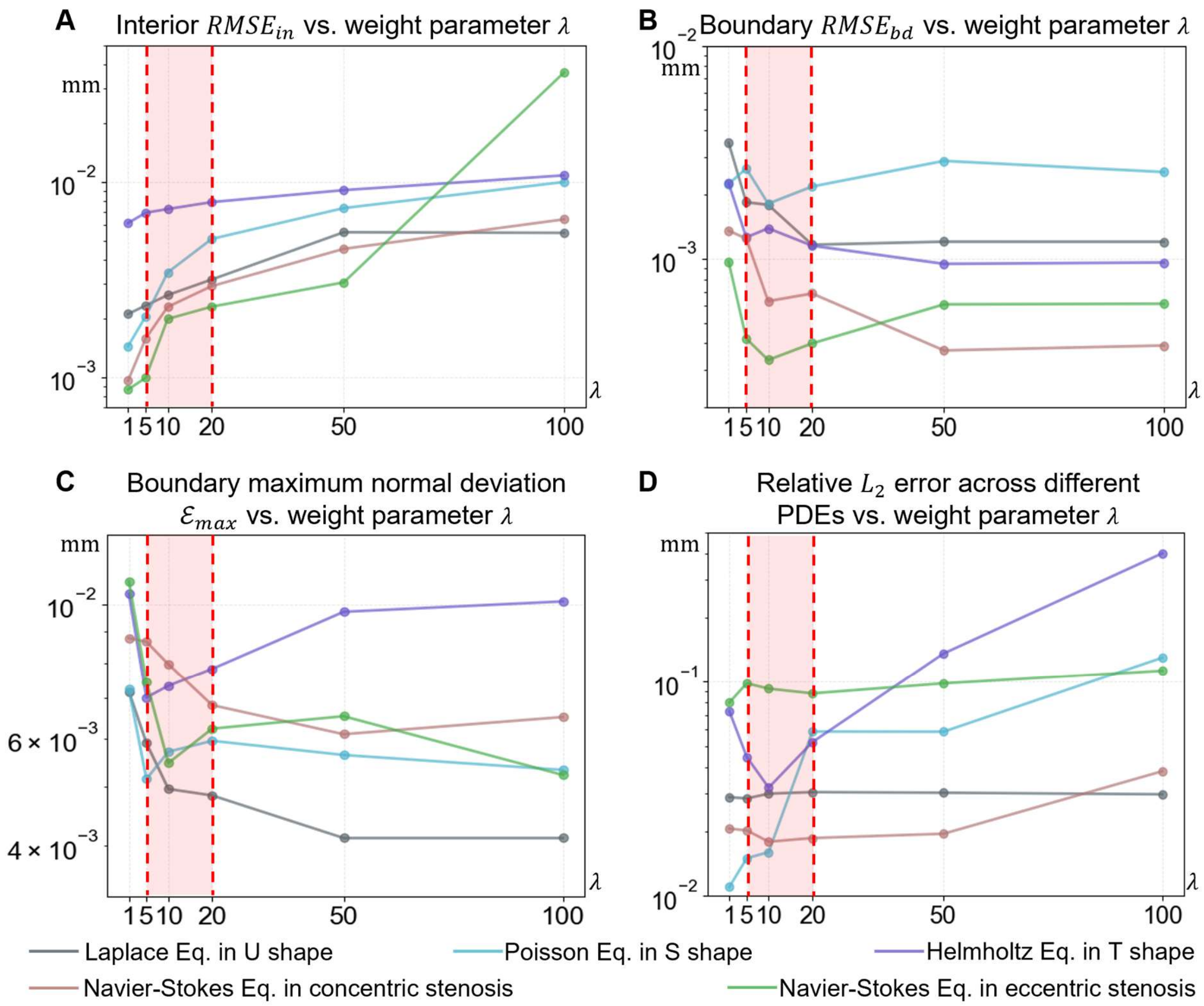

**Figure 9. Sensitivity analysis of boundary weight $\lambda$. (A)** Root mean square error of interior points, $RMSE_{in}$. **(B)** Root mean square error of boundary points, $RMSE_{bd}$. **(C)** Maximum normal deviation of boundary mapped points $\mathcal{E}_{max}$. **(D)** Final PDE prediction relative $L_2$ error. The red bandwidth indicates the optimal range of the weighting coefficient $\lambda \in [5,20]$, within which the model achieves the most stable and accurate performance.

# Appendix-4

### Noise Robustness of the Learned Coordinate Transformation of JacobiNet

To evaluate the robustness of the proposed coordinate transformation under noisy conditions, additional experiments were conducted by injecting Gaussian noise into the normalized coordinates used for training. Specifically, after obtaining the normalized coordinates $(\xi, \eta)$ from the geometric editing operations, zero-mean Gaussian perturbations were added as

$$\tilde{\xi}_i = \xi_i + \epsilon_\xi, \tag{52}$$

$$\tilde{\eta}_i = \eta_i + \epsilon_\eta, \tag{53}$$

where $\epsilon_\xi, \epsilon_\eta \sim \mathcal{N}(0, \sigma^2)$. The noise magnitude is defined relative to the geometric sampling resolution, with the standard deviation given by $\sigma = \alpha h$, where $h$ denotes the sampling spacing and $\alpha$ is a scaling coefficient.

To systematically investigate the effect of noise, four levels of perturbation were considered by varying $\alpha$, corresponding to no noise ($\alpha = 0$) and three noisy settings with increasing noise magnitudes ($\alpha = 0.01, 0.10, 0.30$), which approximately correspond to $1\%$, $10\%$, and $30\%$ noise relative to the sampling resolution. The resulting noisy datasets were then used to train JacobiNet to learn the mapping from the physical coordinates to the normalized coordinates. The transformation quality was evaluated using the positive Jacobian determinant ratio $r_{\det(\mathcal{J})}$, as well as the RMSE on interior and boundary points. In addition, the influence of noise on the downstream physics solution was further analyzed using the relative $L_2$ error.

In the experiments, a non-convex geometry was used as the test case, and the geometric configuration including sampling spacing ($h = 0.05$ m), mapping strategy, and the Poisson equation used for the physics evaluation were kept consistent with those in Section 2.3 of the main text. As shown in Figure 10, the learned mappings remain fold-free under low and moderate noise levels. For the cases with no noise, $1\%$ noise, and $10\%$ noise, the Jacobian determinant ratio remains $r_{\det(\mathcal{J})} = 100\%$, indicating that $\det(\mathcal{J}) > 0$ at all sampled points. When the noise level increases to $30\%$, a small number of negative determinant values appear at a few interior points, resulting in $r_{\det(\mathcal{J})} = 98.64\%$.

Meanwhile, the relative $L_2$ error of the Poisson solution gradually increases with the noise magnitude. Nevertheless, for noise levels up to $10\%$, the error remains acceptable ($L_{2,\mathrm{rel}} < 0.1$). These results indicate that the proposed framework maintains stable and low-distortion coordinate transformations under small to moderate perturbations in the training data.

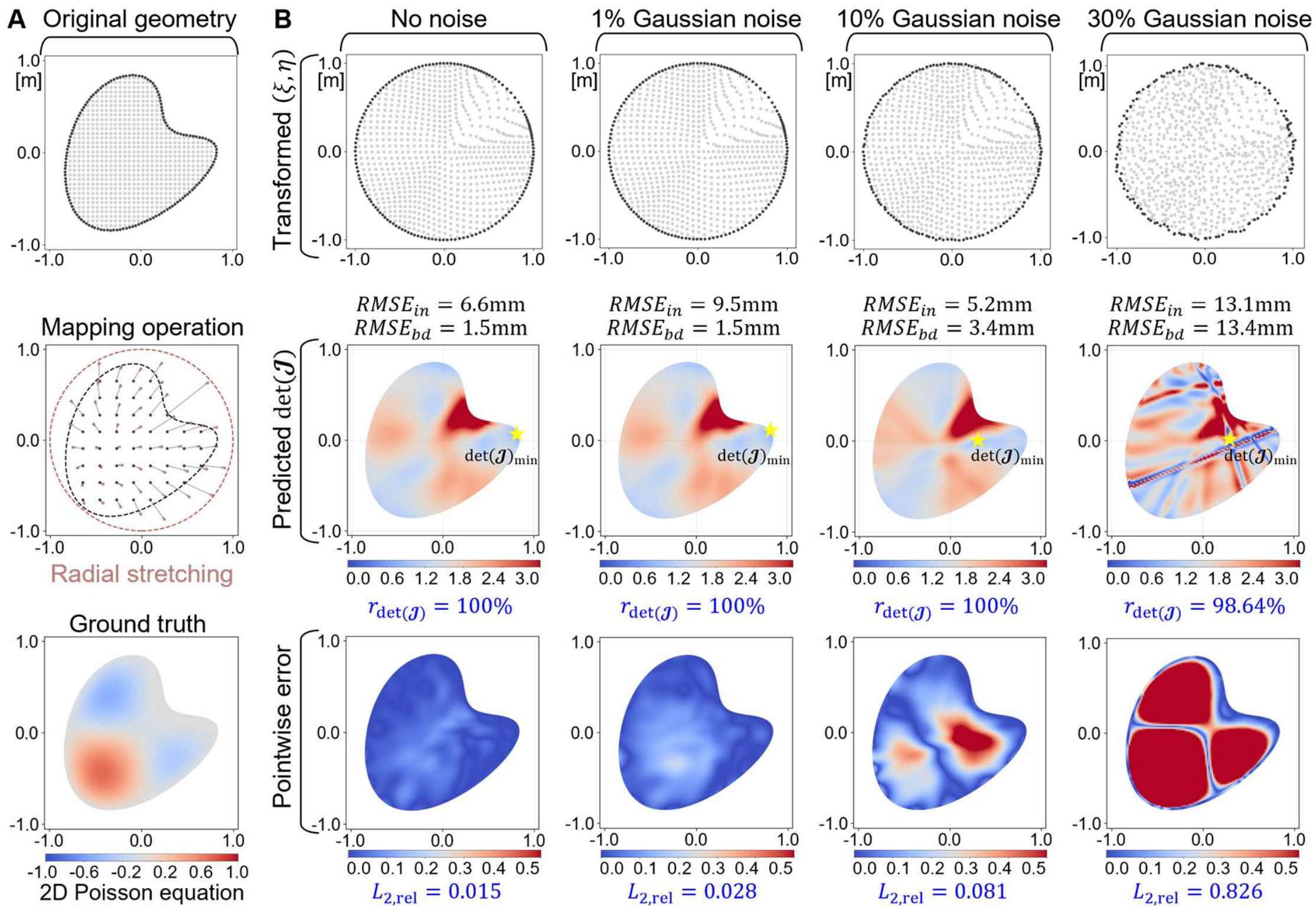

**Figure 10. Noise robustness of the learned coordinate transformation of JacobiNet. (A)** Original geometry, mapping operation, and ground-truth solution of the Poisson equation computed using a high-fidelity FVM solver on a non-convex domain. **(B)** Results under increasing noise levels applied to the normalized coordinates (0%, 1%, 10%, and 30% relative to the sampling resolution). The learned mappings remain fold-free up to 10% noise. At 30% noise, a small region of negative determinants appears, as $r_{\det(\mathcal{J})} = 98.64\%$. The relative $L_2$ error of the Poisson solution increases with the noise magnitude but remains small for noise levels up to 10%.

# Appendix-5

## Experimental Configurations

In this study, JacobiNet is implemented as a fully connected neural network with 2 hidden layers, each consisting of 128 neurons and Tanh activations. Table 5 summarizes the PINN hyperparameters and experimental configurations used in Section 3. For all cases, we adopted uniform sampling. To ensure fairness, all methods were trained under identical configurations for the corresponding test cases.

**Table 5. Detailed experimental configurations**

| Test case | Network depth | Network width | Input→Output dimensions | Epoch (k) | Sampling spacing (mm) | Collocation points |
|---|---|---|---|---|---|---|
| **Laplace Eq. in U shape** | 3 | 64 | 2→1 (x, y)→(u) | 1 | 0.05 | 18,017 |
| **Poisson Eq. in S shape** | 3 | 64 | 2→1 (x, y)→(u) | 3 | 0.05 | 15,499 |
| **Helmholtz Eq. in T shape** | 3 | 64 | 2→1 (x, y)→(u) | 5 | 0.05 | 12,188 |
| **NS Eq. in 3D stenosed vessels (concentric / eccentric)** | 4 | 64 | 3→4 (x, y, z)→(u, v, w, p) | 50 | 0.20 | 20,688 |
| **NS Eq. in 2D parametric vessels (stenosis / aneurysm)** | 3 | 64 | 2→3 (x, y)→(u, v, p) | 10-80 | 0.05 | 10,921/ 24,981 |

For the compared SOTA methods, hyperparameters are selected following the recommended settings in their original works to ensure a fair comparison. Specifically, Grad Norm [25] is implemented with update frequency $f = 1000$ and moving average coefficient $\alpha = 0.9$, while RAR-D [33] adopts $k = 2$ and $c = 0$. The RFF embedding [25, 44] uses a bandwidth parameter $\sigma = 1$, consistent with both the recommended range and configurations used in similar flow problems. For Δ-PINN [50], the number of eigenfunctions is set to $N = 50$ as suggested in the original study. No additional hyperparameter tuning is performed for these methods.

Besides, SiLU is adopted as the activation function in the PINN solver instead of the commonly used Tanh. To validate the rationality of this choice, we conducted comparative experiments to systematically evaluate the performance of several commonly used activation functions, including Tanh, SiLU, ReLU, GELU, and Softplus. As shown in Figure 11 and Table 6, SiLU, GELU, and Tanh perform substantially better than the other activation functions in the baseline PINN. In the JacobiNet–PINN framework, SiLU and GELU further outperform Tanh, indicating improved prediction accuracy and training stability. Similar observations have also been reported in recent studies on activation functions for physics-informed neural networks and PDE learning [57, 58]. It is also important to emphasize that the main conclusions of this work do not depend on the choice of activation function. The introduction of JacobiNet consistently reduces the prediction error across all tested activation functions, demonstrating that the performance improvement brought by JacobiNet is robust with respect to the activation choice.

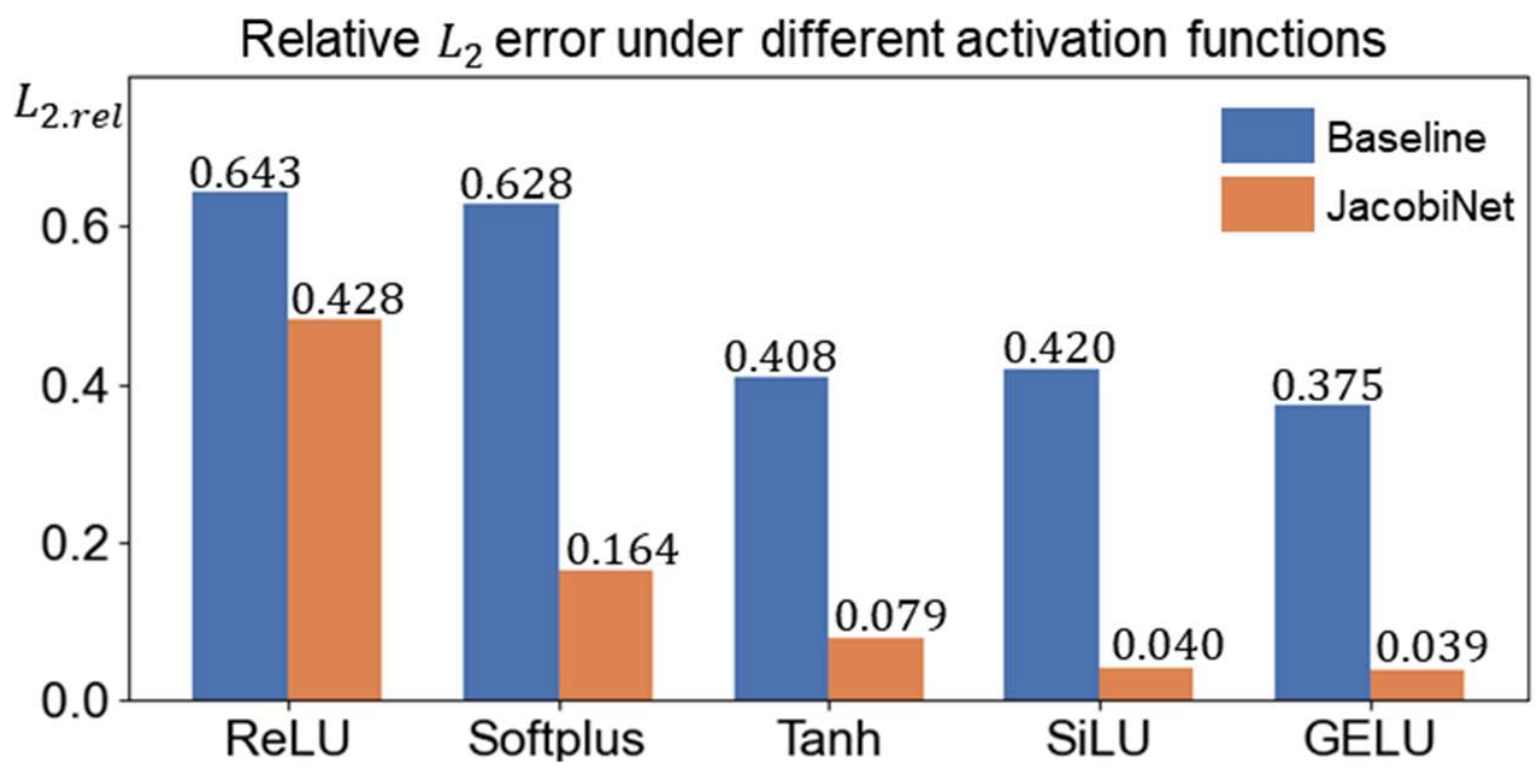

**Figure 11. Relative $L_2$ errors obtained with different activation functions for the baseline and JacobiNet–PINN.** The results indicate that SiLU and GELU tend to yield lower relative $L_2$ errors on average, while JacobiNet consistently improves accuracy across all tested activation functions, indicating robustness to activation choice. For visualization clarity, relative $L_2$ errors exceeding 1 (typically associated with non-convergent cases) are truncated to 1 in the bar chart. This avoids excessively large values dominating the scale and masking the differences among the remaining results. The original numerical values are reported in Table 6.

**Table 6.** Relative $L_2$ error comparison of different activation functions on PDE benchmarks.

| | **Laplace Eq. in U shape** | | **Poisson Eq. in S shape** | | **Helmholtz Eq. in T shape** | | **NS Eq. (Re=500) in 2D stenosis (U/p)** | | **NS Eq. (Re=500) in 2D aneurysm (U/p)** | |
|---|---|---|---|---|---|---|---|---|---|---|
| **Activations** | Baseline | JacobiNet | Baseline | JacobiNet | Baseline | JacobiNet | Baseline | JacobiNet | Baseline | JacobiNet |
| **ReLU** | 0.230 | 0.030 | 0.998 | 0.392 | 307.940 | 1.106 | 0.318/ 0.921 | 0.476/ 1.004 | 0.038/ 0.999 | 0.054/ 0.427 |
| **Softplus** | 0.874 | 0.046 | 0.590 | 0.109 | 3.014 | 0.350 | 0.210/ 0.642 | 0.135/ 0.084 | 0.030/ 1.050 | 0.022/ 0.399 |
| **Tanh** | 0.238 | 0.029 | 0.235 | 0.031 | 2.219 | 0.108 | 0.192/ 0.098 | 0.137/ 0.152 | 0.025/ 1.070 | 0.007/ 0.086 |
| **SiLU** | 0.245 | 0.033 | 0.319 | 0.015 | 4.220 | 0.044 | 0.248/ 0.110 | 0.058/ 0.029 | 0.020/ 1.020 | 0.006/ 0.095 |
| **GELU** | 0.145 | 0.029 | 0.271 | 0.014 | 1.891 | 0.023 | 0.237/ 0.119 | 0.041/ 0.021 | 0.030/ 0.823 | 0.005/ 0.142 |

Additionally, to evaluate the influence of different initialization of network weights and biases on prediction performance, we conducted a robustness analysis with respect to random initialization, as summarized in Table 7. Specifically, two representative 2D Navier–Stokes benchmarks at $Re = 500$ were repeatedly trained using five different random seeds: 1, 5, 10, 50, and 99. The results show that JacobiNet reduced the mean relative $L_2$errors of velocity/pressure from $0.126 \pm 0.103/0.954 \pm 0.973$ for the baseline PINN to $0.024 \pm 0.025/0.071 \pm 0.056$. Notably, JacobiNet not only reduced the average prediction errors but also yielded smaller standard deviations than the baseline PINN, indicating stronger robustness with respect to the random initialization of weights and biases.

**Table 7.** Robustness analysis with respect to random initialization on 2D Navier–Stokes benchmarks at Re=500. Each value denotes the velocity/pressure relative $L_2$ error.

| **2D NS eq.** | **Seeds** | **1** | **5** | **10** | **50** | **99** |
|---|---|---|---|---|---|---|
| **Stenosis (Re=500)** | Baseline | 0.239 / 0.117 | 0.254 / 0.128 | 0.015 / 0.042 | 0.216 / 0.111 | 0.248 / 0.110 |
| | JacobiNet | 0.054 / 0.027 | 0.066 / 0.032 | 0.013 / 0.015 | 0.014 / 0.010 | 0.058 / 0.029 |
| **Aneurysm (Re=500)** | Baseline | 0.131 / 1.908 | 0.045 / 2.440 | 0.042 / 1.490 | 0.048 / 2.170 | 0.020 / 1.020 |
| | JacobiNet | 0.005 / 0.168 | 0.007 / 0.135 | 0.006 / 0.109 | 0.006 / 0.093 | 0.006 / 0.095 |

# Appendix-6

## Paired Data Generation for JacobiNet

JacobiNet is trained using paired point correspondences. For simple geometries, such correspondences can be obtained via analytical or diffeomorphic mappings. For more complex domains, where conventional parameterizations are unavailable, we employ geometry editing operations such as filling, radial stretching, and cut-and-unfold transformations to generate point-wise mappings. It is worth noting that these geometry editing operations typically cannot provide the Jacobian of the mappings, as they involve non-differentiable steps (e.g., nearest-neighbor search and discrete indexing), which prevents the mappings from being embedded into end-to-end differentiable computational graphs and thus limits their use in physics-based coordinate transformations. However, the key advantage of JacobiNet—learning a differentiable mapping with an implicitly embedded Jacobian—holds irrespective of how the paired data are constructed.

In the following, we detail the two representative paired data generation algorithms used in this work.

### 1. Radial stretching operation

For non-convex geometries, paired data can be constructed using a *radial stretching* strategy.

**Algorithm 1 Radial Stretching Operation**

1. **Input:** Boundary points $\{\mathbf{x}_k\} \subset \mathbb{R}^2$, sampled points $\mathcal{X} = \{\mathbf{x}_m\} \subset \mathbb{R}^2$.
2. Compute the geometric center:
$$\mathbf{c} = \frac{1}{N_b}\sum_{k=1}^{N_b} \mathbf{x}_k.$$
3. Construct unit boundary directions:
   **for** $k = 1,2,\dots,N_b$ **do**
$$\mathbf{u}_k = \frac{\mathbf{x}_k - \mathbf{c}}{\| \mathbf{x}_k - \mathbf{c} \|}.$$
   **end for**
4. Build a KD-tree using the set of directions $\{\mathbf{u}_k\}$.
5. **For** each point $\mathbf{x}_m \in \mathcal{X}$ **do**
   (a) Compute the direction vector:
$$\mathbf{v}_m = \mathbf{x}_m - \mathbf{c}, \mathbf{u}_m = \frac{\mathbf{v}_m}{\| \mathbf{v}_m \|}.$$
   (b) Query the KD-tree to find the nearest boundary direction $k^*$.
   (c) Compute the corresponding radius:
$$R = \| \mathbf{x}_{k^*} - \mathbf{c} \|.$$
   (d) Normalize:
$$(\xi_m, \eta_m) = \frac{\mathbf{v}_m}{R}.$$
   **end for**
6. **Output:** Normalized coordinates $\{(\xi_m, \eta_m)\}$.

Conceptually, this process can be viewed as a non-uniform normalization along directions, where the scaling factor is geometry dependent. In this work, it is applied to the Hexagram (Fig. 2) and the non-convex geometry (Fig. 3), while the L-shape filling (Fig. 2) and Y-slice min-max norm of T-shape (Fig. 4) can be interpreted as a generalized extension under the same geometric editing framework.

**2. Unfolding along centerline**

For geometries with a dominant direction (e.g., curved pipes or vascular structures), paired data can be constructed using an *unfolding along centerline* strategy.

**Algorithm 2 Unfolding Along Centerline Operation**

1. **Input:** Centerline points $\{\mathbf{c}_k\} \subset \mathbb{R}^2$, sampled points $\mathcal{X} = \{\mathbf{x}_m\} \subset \mathbb{R}^2$.
2. Parameterize the centerline by arc length:
$$s_k = \sum_{j=1}^{k} \| \mathbf{c}_j - \mathbf{c}_{j-1} \|.$$
3. Compute tangent and normal vectors along the centerline:
**for** $k = 1,2,\ldots,N_c$ **do**
$$\mathbf{t}_k = \frac{d\mathbf{c}}{ds}|_{s=s_k}, \mathbf{n}_k = (-t_{k,y}, t_{k,x}).$$
**end for**
4. **For** each point $\mathbf{x}_m \in \mathcal{X}$ **do**
(a) Find the nearest sampled centerline point:
$$k^* = \arg\min_r \| \mathbf{x}_m - \mathbf{c}_r \|.$$
(b) Assign arc-length coordinate:
$$\tilde{\eta}_m = s_{k^*}.$$
(c) Compute relative displacement:
$$\mathbf{v}_m = \mathbf{x}_m - \mathbf{c}_{k^*}.$$
(d) Project onto normal direction:
$$\tilde{\xi}_m = \mathbf{v}_m \cdot \mathbf{n}_{k^*}.$$
**end for**
5. Normalize the coordinates:
$$\xi_m = 2 \cdot \frac{\tilde{\xi}_m - \tilde{\xi}_{\min}}{\tilde{\xi}_{\max} - \tilde{\xi}_{\min}} - 1, \eta_m = 2 \cdot \frac{\tilde{\eta}_m - \tilde{\eta}_{\min}}{\tilde{\eta}_{\max} - \tilde{\eta}_{\min}} - 1.$$
6. **Output:** Normalized coordinates $\{(\xi_m, \eta_m)\}$.

In this work, it is applied to tubular geometries (e.g., the "U" and "S" shapes in Fig. 4), enabling complex curved domains to be mapped to regular rectangular reference domains.

Although JacobiNet enables geometry editing operations with differentiable Jacobians, thereby expanding the range of geometries amenable to coordinate transformations, mappings for highly complex domains remain unknown. Generating supervised paired data for highly complex geometries, as discussed in Section 4, such as branching vessels, remains an open challenge due to the difficulty of constructing consistent point-wise correspondences.